\documentclass[10pt,twocolumn,letterpaper]{article}
\usepackage{cvpr}              

\definecolor{cvprblue}{rgb}{0.21,0.49,0.74}
\usepackage[pagebackref,breaklinks,colorlinks,allcolors=cvprblue]{hyperref}

\usepackage{pifont}
\usepackage{graphicx}
\usepackage{xcolor}
\usepackage{colortbl, booktabs}
\usepackage{multirow}
\usepackage{bbding}
\usepackage{amsmath}
\usepackage{amsfonts}
\usepackage{bm}

\newcommand\blfootnote[1]{%
  \begingroup
  \renewcommand\thefootnote{}\footnote{#1}%
  \addtocounter{footnote}{-1}%
  \endgroup
}

\usepackage{subcaption}
\definecolor{cred}{HTML}{FF6B6B}
\definecolor{corange}{HTML}{FF5200}
\definecolor{cgreen}{HTML}{70AD47}
\definecolor{cblue}{HTML}{686EE2}
\definecolor{cpurple}{HTML}{A149FA}
\definecolor{ggray}{RGB}{127,127,127}
\definecolor{aliceblue}{rgb}{0.94, 0.97, 1.0}
            
\def\paperID{9566} 
\def\confName{CVPR}
\def\confYear{2025}

\title{NeighborRetr: Balancing Hub Centrality in
Cross-Modal Retrieval}

\author{
    Zengrong Lin$^{1^*}$\qquad
    Zheng Wang$^{1^* \dagger}$\qquad
    Tianwen Qian$^{2}$\qquad
    Pan Mu$^{1}$\qquad
    Sixian Chan$^{1}$\qquad
    Cong Bai$^{1}$
    \\
    \{linzengrong, zhengwang, panmu, sxchan congbai\}@zjut.edu.cn, twqian@cs.ecnu.edu.cn\\
    $^{1}$College of Computer Science and Technology, Zhejiang University of Technology, Zhejiang, China \\
    $^{2}$College of Computer Science and Technology, East China Normal University, Shanghai, China
    \\
}

\begin{document}
\maketitle
\begin{abstract}

\blfootnote{%
  \begin{tabular}[t]{@{}l@{}}
    $^*$ Equal Contribution.\\
    $^\dagger$ Corresponding Author: Zheng Wang $<$zhengwang@zjut.edu.cn$>$.
  \end{tabular}%
}
Cross-modal retrieval aims to bridge the semantic gap between different modalities, such as visual and textual data, enabling accurate retrieval across them. Despite significant advancements with models like CLIP that align cross-modal representations, a persistent challenge remains: the hubness problem, where a small subset of samples (hubs) dominate as nearest neighbors, leading to biased representations and degraded retrieval accuracy. Existing methods often mitigate hubness through post-hoc normalization techniques, relying on prior data distributions that may not be practical in real-world scenarios. In this paper, we directly mitigate hubness during training and introduce NeighborRetr, a novel method that effectively balances the learning of hubs and adaptively adjusts the relations of various kinds of neighbors. Our approach not only mitigates the hubness problem but also enhances retrieval performance, achieving state-of-the-art results on multiple cross-modal retrieval benchmarks. Furthermore, NeighborRetr demonstrates robust generalization to new domains with substantial distribution shifts, highlighting its effectiveness in real-world applications.
We make our code publicly available at: \href{https://github.com/zzezze/NeighborRetr}{https://github.com/zzezze/NeighborRetr}.


\end{abstract}

\section{Introduction}
\label{sec:intro}

Cross-modal retrieval has long been the touchstone of cross-modal representation learning, with the primary goal of bridging the semantic gap between different modalities for accurate retrieval.
In recent years, visual-language models such as CLIP~\cite{radford2021learning} have gained significant attention due to their remarkable performance in aligning cross-modal representations. By leveraging large-scale paired datasets, these models enable contrastive learning to align visual and textual data distributions, which has notably advanced cross-modal retrieval tasks.

\begin{figure}[t]
\centering
\includegraphics[width=1.0\linewidth]{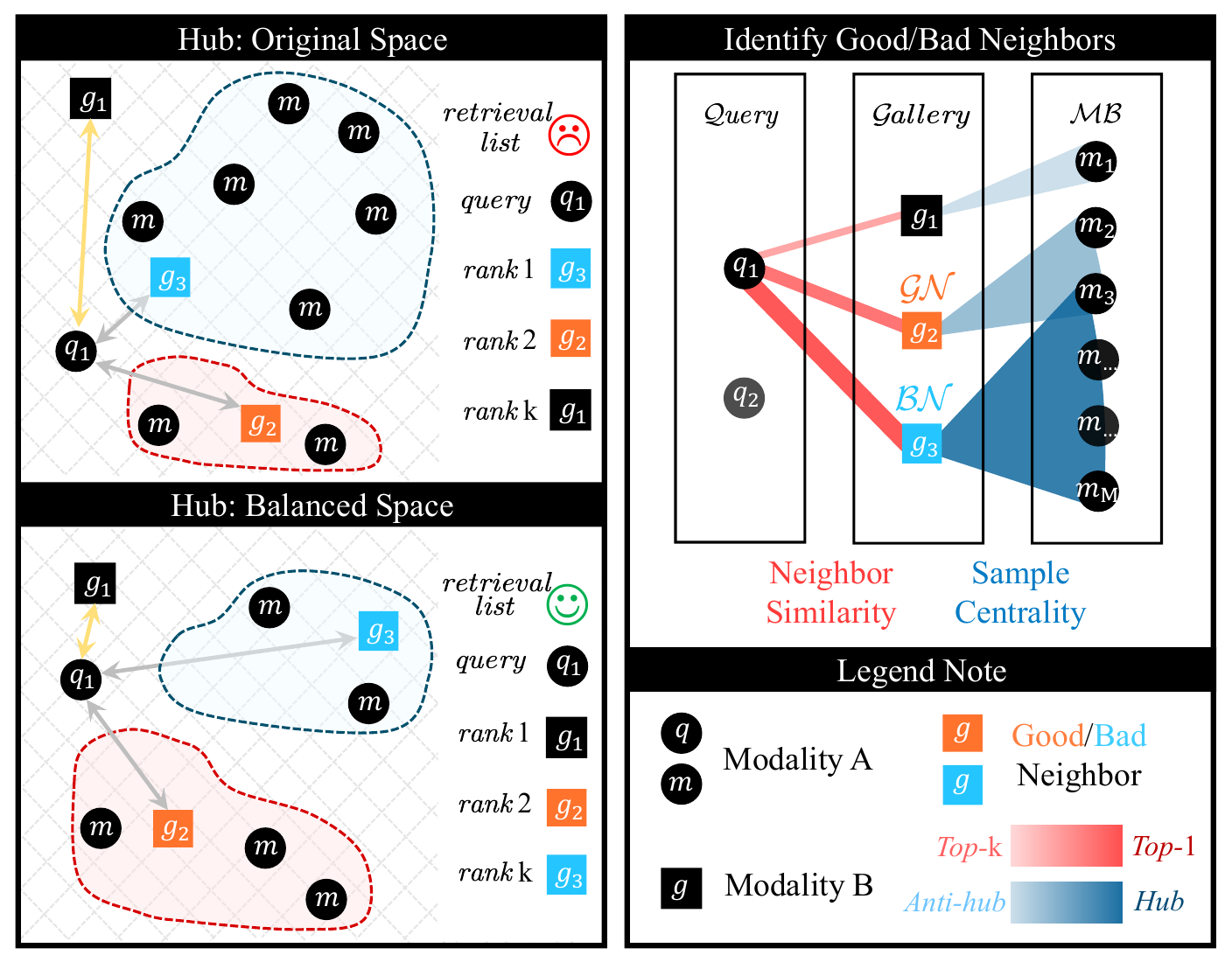}
\caption{
\emph{Left:} \textbf{Hubness Balancing.} In the original embedding space (top left), bad hubs such as $g_3$ dominate the neighborhood, leading to unsatisfying retrieval performance. NeighborRetr rebalances the neighborhood relations  (bottom left) by adaptively bringing good hubs closer, such as $g_2$, effectively mitigating the hubness problem.
\emph{Right:} \textbf{Good/Bad Neighbors Identification.} NeighborRetr identifies cross-modality hubs based on sample centrality with Memory bank (MB), and distinguishes between good neighbors (GN) and bad neighbors (BN) by considering both the neighbor similarity and the sample centrality. 
}
\label{fig:head}
\vspace{-1.2em}
\end{figure}


However, despite these advancements, a significant challenge remains: hubness. Hubness is a common phenomenon in high-dimensional embedding spaces~\cite{tomavsev2015hubness}, where a small subset of samples, referred to as \emph{hubs}, frequently emerge as the nearest neighbors to other samples, while the majority of other samples rarely appear as neighbors. As illustrated in the top left of Fig.~\ref{fig:head}, hubs such as $g_3$ dominate the nearest-neighbor rankings, leading to biased representations and degraded retrieval accuracy.


Even large-scale pretrained models like CLIP are not immune to the hubness problem~\cite{bogolin2022cross}. Recently, some methods have attempted to mitigate hubness by applying post-hoc similarity normalization techniques during inference. These methods either leverage the distribution of the test set~\cite{cheng2021improving} or construct additional modality-specific banks from training samples~\cite{bogolin2022cross,wang2023balance}.
Lately, NNN~\cite{chowdhury2024nearest} normalizes the retrieval score using the k-closest queries from a reference dataset.
However, these approaches rely heavily on knowing the prior data distribution of either the training or test sets, which can be problematic in real-world scenarios where the test set distribution may shift or remain unknown. 
This limitation highlights the importance of addressing hubness during training to ensure robust performance across varying distributions.
To tackle hubness during training, earlier methods, such as HAL~\cite{liu2020hal}, scale up the loss of hubs of both positive and negative samples. 
However, this strategy does not differentiate between semantically relevant and irrelevant hubs, potentially punishing beneficial hubs.
A more nuanced approach that distinguishes between relevant (good) hubs and irrelevant (bad) hubs is essential for effectively mitigating the hubness problem while maximizing retrieval accuracy.


To achieve nuanced hubness reduction during training, we empirically show in Sec.~\ref{sec:hub} and Fig.~\ref{fig:hub_bc}(a) that although semantical relevant hubs and irrelevant hubs co-exist, while relevant hubs exhibit lower retrieval frequency than irrelevant ones, making it possible to distinguish them.
Inspired by this, we propose NeighborRetr, a novel method that aims to balance the centrality~\cite{hara2016flattening} of hubs in the embedding space. Rather than categorically rejecting all hubs, we introduce sample centrality for identifying hubs and further distinguish between good neighbors (semantically relevant samples) and bad neighbors (irrelevant or noisy samples) incorporating neighbor similarity with sample centrality. By selectively adjusting the affinity strength with all neighbors, NeighborRetr rebalances the neighbor relationships to bring good hubs closer than bad ones. Additionally, we leverage the uniform retrieving objective to further enforce the retrieval of anti-hubs.


Our main contributions are as follows: 
(i) We conduct an in-depth investigation into the cross-modal hubness problem, revealing the presence of both beneficial good hubs and detrimental bad hubs, as well as overlooked anti-hubs.
(ii) We introduce NeighborRetr, a novel method that effectively balances the learning of hubs and adjusts neighborhood relationships to mitigate hubness.
(iii) Our approach not only mitigates the hubness problem but also enhances retrieval performance across multiple benchmarks, achieving state-of-the-art results on four text-video benchmarks and three text-image benchmarks. In addition, our method demonstrates robust generalization to new domains with substantial distribution shifts.

\section{Related Works}

\noindent\textbf{Cross-Modal Matching.}\quad
Triplet ranking loss~\cite{kiros2014unifying} takes the sum of all negative samples for visual semantic learning.
VSE++~\cite{faghri2017vse++} emphasizes hard negatives and contrasts the hardest negative sample.
Wray et al.~\cite{wray2021semantic} argue that visual similarity does not imply semantic similarity, and propose several proxies measuring text similarity to allow relative ranking. 
PVSE~\cite{song2019polysemous} represents each sample as a set of embedding vectors. 
PCME~\cite{chun2021probabilistic} introduces probabilistic embedding representing each sample as a set of vectors sampled from a normal distribution.
Li et al.~\cite{li2023integrating} utilize language pre-training models to identify semantic similarity and adjust the margin of false negatives.
Kim et al.~\cite{kim2023improving} regard semantic alignment as a set prediction problem and propose the smooth-Chamfer similarity to address the set alignment problem.
MSRM~\cite{wang2023multilateral} introduces hypergraph into the one-to-many correspondence of image-text retrieval. 

\noindent\textbf{Hubness Problem.}\quad
The hubness is a fundamental property observed in high-dimensional representation. 
Jian et al.~\cite{jian2023invgc} highlight representation degeneration in multi-modal data, hindering retrieval performance.
It has been empirically observed that hubness also exists for cross-modal matching. 
Liu et al.~\cite{liu2019strong} looked into the hubness problem in cross-modal matching, and proposed the kNN-margin loss that considers all k-hardest samples as negatives. 
Furthermore, HAL~\cite{liu2020hal} alleviates the hubness problem by using global measurements for hubness level and scaling up hub losses during learning.
In addition to training adjustments, recently, more methods handle hubness during inference.
Dual Softmax~\cite{cheng2021improving} performs prior normalization in both directions as a postprocessing step. 
QB-NORM~\cite{bogolin2022cross} reduces the prominence of the hub with the query bank built from queries in the training set.
DBNORM~\cite{wang2023balance} leverages two banks of query and gallery samples from the training set to reduce hubness during inference.

\noindent\textbf{Visual-Text Retrieval.}\quad
ViLT~\cite{kim2021vilt} introduces a minimal vision-and-language transformer, simplifying visual input processing.
SGRAF~\cite{diao2021similarity} further enhances image-text retrieval by effectively capturing and filtering both local and global alignments.
To enhance video representations for text-video retrieval, 
approaches like CLIP4Clip~\cite{luo2021clip4clip} have sought to overcome this obstacle by leveraging knowledge transferred from CLIP~\cite{radford2021learning}, while still regards videos as a whole representation. 
X-Pool~{\cite{gorti2022x}} allows a text to focus on its most semantically similar video frames, subsequently creating an aggregated video representation conditioning on these specific frames.
TS2-Net~{\cite{liu2022ts2}} introduces the token shift and selection transformer, enhancing the temporal and spatial video representations.
To enable fine-grained cross-modal matching for text-video retrieval,
UATVR~\cite{fang2023uatvr} allows for both global and local probabilistic alignment between text and video feature representations.
HBI~\cite{jin2023video} incorporates Hierarchical Banzhaf Interaction to value possible correspondence between video frames and text words.
DiCoSA~{\cite{jin2023text}} aligns text and video via semantic concepts, using adaptive pooling for set-to-set and partial matching.
To regularize semantic samples, 
Tomas et al.~\cite{thomas2022emphasizing} propose a novel approach to prioritize loosely aligned samples.

\section{Problem Setup}
\subsection{Preliminary}

\noindent\textbf{Task definition.}\quad
Generally, given a corpus of visual-text pairs $(v, t)$, cross-modal representation learning aims to learn a visual encoder $\Phi_v$ and a text encoder $\Phi_t$. 
The optimization target is formulated as a cross-modality similarity measurement $S(\Phi_v(v),\Phi_t(t))$ by contrasting cross-modal samples, where maximize the similarity for matched pairs and minimize it for mismatched pairs. 
For cross-modal retrieval, a query $x $ from one modality is compared with gallery $y$ from another modality in the joint embedding space. The goal is to generate a ranked list of galleries that best match the query $x$.

\noindent\textbf{Hubness Problem.}\quad
In a dataset ${\mathcal{X}} \subset \mathbb{R}^{d}$, the affinity between a sample $x\in \mathcal{X}$ with other samples ${x'}\in \mathcal{X}$ can be measured by inner product similarity, with the nearest neighbors being those with the maximum similarities. 
The k-occurrence of $x$, denoted as $N_{k}(x)$, is the number of times $x$ appears among the k-nearest neighbors of other samples.  A sample is considered a \emph{hub} when $N_{k}(x)$ is significantly higher than others.
It further leads to the phenomenon known as \emph{hubness}, where a small portion of samples become hubs.
As described by Radovanovic et al.~\cite{radovanovic2010hubs}, the distribution of $N_{k}(\mathcal{X})$ often exhibits high skewness, indicating the presence of hubness.
Oppositing to hubs, \emph{anti-hubs} are those samples that appear in very few neighbors of other samples.

\noindent\textbf{Good/Bad Hubs.}\quad
In the presence of class labels, Tomavsev et al.~\cite{tomavsev2015hubness} introduce the concepts of \emph{good hub} and \emph{bad hub}.
Specifically, $x$ is a \emph{good (or bad) neighbor} of another sample~$x'$ if: (i)~$x$ is among the $k$-nearest neighbors of $x'$, and (ii)~$x$ and $x'$ have the same (or different) class label. The good (or bad) $k$-occurrence of $x$, denoted as $GN_{k}(x)$ (or $BN_{k}(x)$), counts how many other samples have $x$ as one of their good (or bad) $k$-nearest neighbors. A sample $x$ is a \emph{good hub} (or \emph{bad hub}) if $GN_{k}(x)$ (or $BN_{k}(x)$) is exceptionally large.
\begin{figure}[t]
\centering
\includegraphics[width=1.0\linewidth]{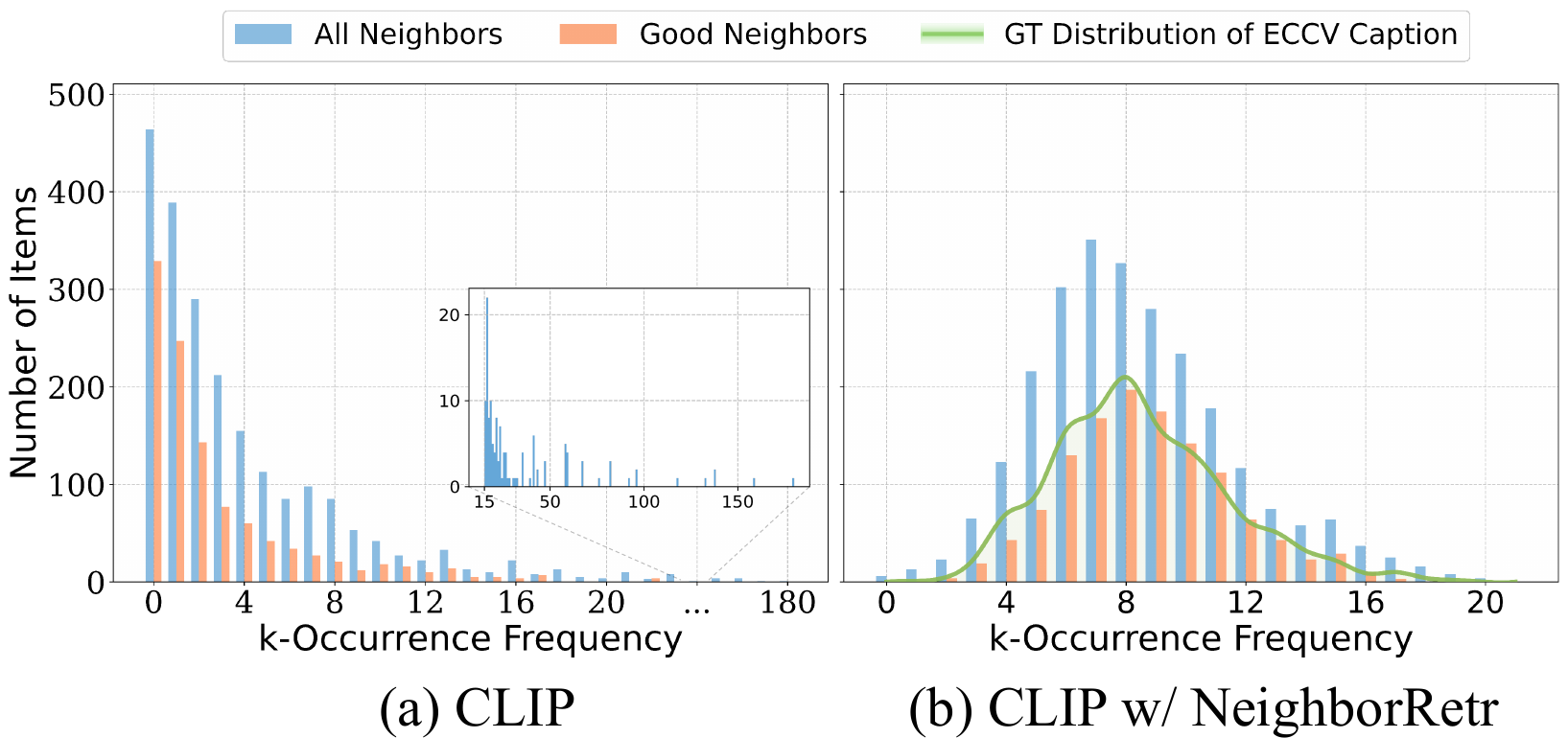}
\caption{Distribution of k-occurrence frequency among all textual similarities for
\textbf{(a) vanilla CLIP} and \textbf{(b) CLIP fintuned with NeighborRetr}.  
The x-axis represents how often an image appears in queries’ top-15 nearest neighbors. The y-axis counts the number of images at each frequency.
Blue bars indicate all neighbors, orange bars highlight positive pairs among neighbors, and the green line represents the ground truth distribution.
}
\label{fig:hub_bc}
\vspace{-1.0em}
\end{figure}
\subsection{Observation and Analysis}
\label{sec:hub}
\noindent\textbf{Hubness Observation.}\quad
In typical cross-modal retrieval tasks, each sample has a single labeled positive counterpart, with all other counterparts treated as negatives. This setup contrasts with real-world scenarios where multiple positive counterparts may exist.
To categorize unlabeled cross-modal pairs as pseudo-positive or negative, we propose using intra-text similarity as a probe to identify potential cross-modal multi-correlations. Concretely, we use CLIP’s text encoder to obtain the representations of the text originally paired with the image and consider these with high textual similarity as additional positives 
(further rationale for this is provided in the supplementary material). Then, to investigate the hubness issue of cross-modal embedding space, we compare cross-modal similarities across all unlabeled samples and count how often each image appears among the top 15 nearest neighbors (k-occurrence) of text queries in the test set of ECCV Caption~\cite{chun2022eccv}.
The results are presented in Fig.~\ref{fig:hub_bc} and analyzed below.

\noindent\textbf{Hubness Analysis.}\quad
In Fig.\ref{fig:hub_bc} (a),  we show the k-occurrence frequency of text-to-image retrival. From this plot, we draw three observations: (i) The distribution of all neighbors exhibits an extreme long-tail pattern, with a high presence of bad hubs (undesirable large $N_{k}(x)$), as highlighted in the inset. (ii) Good neighbors are distributed across a range of lower $N_{k}(x)$ frequencies. (iii) Many anti-hubs (with zero or very small $N_{k}(x)$) seldom appear among the top-15 nearest neighbors of any query. 

To address these issues, rather than indiscriminately eliminating all hubs and ignoring anti-hubs,
our proposed NeighborRetr method mitigates the hubness problem in multiple aspects. 
In Fig.\ref{fig:hub_bc} (b), we analyze the CLIP model fine-tuned with NeighborRetr, showing that (i) bad hubs are reduced, (ii) good hubs are enhanced, and (iii) anti-hubs are minimized. These changes lead to a distribution of good neighbors that better aligns with the ground truth. This indicates that NeighborRetr creates a more balanced embedding space, capturing true semantic correlations of ECCV Caption. In the following sections, we introduce the method of NeighborRetr and further explain how it achieves such changes.

\section{Methodology}
\subsection{Sample Centrality}
Centrality, introduced in~\cite{radovanovic2010hubs}, refers to the tendency of samples closer to the global centroid to exhibit greater affinity to all other samples.
Hubness can be understood as a kind of local centrality~\cite{tomavsev2014role}, reflecting a localized centering phenomenon.
Since hubs frequently appear among the k-nearest neighbors of other samples,  we define a \emph{centrality score} to measure a sample's affinity with others. 
Directly calculating the centrality score across all training set samples can be computationally intensive, we manage this by maintaining a queue~\cite{he2020momentum} as a memory bank $\mathcal{M}$ to efficiently compute the uni-modality centrality score:
\begin{align}
    C(x_{i})=\frac{1}{|\mathcal{M}|}\sum_{j=1}^{|\mathcal{M}|}{\frac{x_{i}^{\top}x_{j}}{||x_{i}||\cdot||x_{j}||}},
    \label{eq:centrality}
\end{align}
where  $|\mathcal{M}|$ is the size of the memory bank. $C(x_{i})$ measures the average similarity of $x_i$ to all memory samples $x_j $. 
A high centrality score suggests a strong global affinity.
  
Similarly, in the joint embedding space, the centrality score for a sample $y_{j}$ in one modality can be defined as its average similarity with all samples $x_{i}$ in another modality:
\begin{align}
    \hat{C}(y_{j})=\frac{1}{|\mathcal{M}|}\sum_{i=1}^{|\mathcal{M}|}{\frac{y_{j}^{\top}x_{i}}{||y_{j}||\cdot||x_{i}||}}.
\end{align}
\vspace{-0.0em} 
\subsection{Centrality Weighting}
Previous methods for mitigating cross-modal hubness~\cite{liu2020hal} apply inter-modality centrality scores for softly weighting various hubs. Here, we propose to weight hubs leveraging centrality within the intra-modality similarity. We define the centrality weight as:
\begin{align}
    w(x_{i})=\exp(C(x_{i})/\kappa),
\end{align}
where $C(x_{i})$ is the uni-modal centrality score as in Eq.~\ref{eq:centrality}, and $\kappa$ is a scaling hyperparameter. 
To emphasize the learning of hubs, we incorporate the centrality weight into the contrastive loss, defining the {Centrality Weighting Loss} as:
\begin{equation}
\mathcal{L}_{{Wti}}= -\frac{1}{B}\sum_{i=1}^{B}w(x_{i})\log\frac{\exp(\textrm{S}(x_i,y_i))}{\sum_{j}^{B}\exp(\textrm{S}(x_i,y_j))}.
\label{eq:hcw}
\end{equation}
where $B$ is the batch size, and $S$ is the cross-modal similarity. 
Here, we weight the contrastive loss with intra-modal centrality scores instead of cross-modal centrality scores because intra-modality scores provide a more stable representation of hub within each modality, enabling precise hub identification, whereas cross-modal scores may conflate
interdependencies within different modalities and obscure modality-specific hub centrality.  
By incorporating intra-modal weights into the loss, we can accurately identify hubs with high centrality scores and emphasize the centrality of hubs within each modality.
\vspace{0.5em} 
\subsection{Neighbor Adjusting}
During contrastive learning, it is crucial to handle hubs effectively by bringing good hubs closer to the anchor sample and distancing it from bad ones. Ideally, if multi-correlations are available, a supervised contrastive learning function can be employed. However, in most cases, each sample has only a single labeled positive counterpart. In addition, the cross-modal similarity measure is often biased in positive-negative discrimination due to the hubness issue within the embedding space.  Moreover, centrality scores alone can only identify whether a sample is a hub, without distinguishing between good and bad hubs.
To address this, we incorporate the centrality score into the cross-modal similarity measure, rewriting it as:
\begin{align}
    \tilde{\textrm{S}}(x_{i}, y_{j}) = \textrm{S}(x_{i}, y_{j}) - \hat{C}(y_{j}).
\end{align}
With this de-centrality cross-modal similarity, we focus on balancing contrastive neighbors within data batches.
Specifically, for a query $x_{i}$, we define a set of gallery samples $\mathcal{N}(x_{i})$ as its closest neighbors. 
For convenience, we also define the neighbor set including the ground truth $y_{i}$:
    $\mathcal{N}^+(x_i) = \{ y_i \} \cup \mathcal{N}(x_i)$.
To differentiates various kinds of affinity in the neighborhood, we introduce the Neighbor Adjusting Loss $\mathcal{L}_{\text{Nbi}}$ as:
\begin{align}
       \mathcal{L}_{{Nbi}}(x_{i}) = - \sum_{y_j \in \mathcal{N}^+(x_i)} \mathcal{H}(y_{j})
    \log P(y_j|\mathcal{N}^+(x_i)),
    \label{eq:hub_loss}
\end{align}
where 
\vspace{-0.5em}
\begin{equation}
  \begin{aligned}
P(y_j|\mathcal{N}^+(x_i)) = \frac{\exp(\textrm{S}(x_i,y_j))}{\sum_{k=1}^{|\mathcal{N}^+|} \exp(\textrm{S}(x_i,y_k))},
  \end{aligned}
  \label{eq:pni}
\end{equation}
and
\begin{equation}
    \mathcal{H}(y_{j}) = \dfrac{\exp\left( \tilde{\textrm{S}}(x_{i}, y_{j}) \right)}
    {\displaystyle \sum\limits_{y_{k} \in \mathcal{N}(x_{i})} \exp\left( \tilde{\textrm{S}}(x_{i}, y_{k}) \right)},
\end{equation}
and $\mathcal{H}(y_j) = 1.0$ if $y_j = y_{i}$.
Here, $P(y_j|\mathcal{N}^+(x_{i}))$ quantifies the likelihood of $x_i$ matching $y_j$ among neighbors, normalized over $\mathcal{N}^+(x_{i})$ using their similarity scores.
$\mathcal{H}(y_{j})$ adaptively adjusts the contribution of each neighbor based on $\tilde{\textrm{S}}(x_{i}, y_{j})$. Intuitively, $\mathcal{H}(y_{j})$ penalizes bad hubs with high centrality while promoting good hubs.
And the gradient of $\mathcal{L}_{\text{Nbi}}$ with respect to the similarity scores $\mathbf{S}_{i,j}$ is computed as:
\begin{equation}
\begin{aligned}
    \frac{\partial \mathcal{L}_{{Nbi}}(x_i)}{\partial \textrm{S}(x_i, y_j)} = P(y_j | \mathcal{N}^+(x_i)) - \mathcal{H}(y_j).
\end{aligned}
\label{eq:gradient_nbi}
\end{equation}
The gradient adaptively adjusts similarity scores to promote good hubs and penalize bad ones. Specifically, when $ P < \mathcal{H}(y_j) $ (good hub), the negative gradient increases $ S(x_i, y_j) $, promoting the hub. Conversely, when $ P > \mathcal{H}(y_j) $ (bad hub), the positive gradient decreases $ S(x_i, y_j) $, penalizing the hub. This adaptively balances the representation of good and bad hubs.

\vspace{-0.0em} 
{\subsection{Uniform Regularization}}

While $\mathcal{L}_{\text{Wti}}$ and $\mathcal{L}_{\text{Nbi}}$ losses mitigate hubness by targeting hubs, they overlook anti-hubs that are rarely or never retrieved. To address the anti-hub issue~\cite{tomasev2011probabilistic}, inspired by~\cite{distances2013lightspeed},
we propose a uniform marginal constraint that treats all samples with equal probability of retrieval, ensuring anti-hubs have retrieval probabilities comparable to normal samples. This uniform retrieving objective enforces balanced retrieval probabilities across all samples:
\vspace{-0.7em}
\begin{equation}
\begin{split}
      \max_{\mathbf{Q} \in \mathbb{R}_{+}^{n \times m}} & \quad 
      \langle \mathbf{Q},\, \mathbf{S} \rangle = \sum_{i=1}^{n} \sum_{j=1}^{m} \mathbf{Q}_{i,j} \mathbf{S}_{i,j}, \\
      \text{s.t.} \quad &  
      \mathbf{Q} \mathbf{1}_m=\frac{1}{n} \mathbf{1}_n, \quad 
      \mathbf{Q}^{\top} \mathbf{1}_n= \frac{1}{m} \mathbf{1}_m,
\end{split}
\label{eq:fair_matching}
\end{equation}
where $\langle \mathbf{Q},\, \mathbf{S} \rangle$ denotes the Frobenius inner product, and $n$, $m$ are sizes of the query and gallery sets, respectively. 
The constraints $\frac{1}{n} \mathbf{1}_n$ and $\frac{1}{m} \mathbf{1}_m$ enforce uniform marginal distributions, ensuring bidirectional retrieval by equally matching each query with all gallery samples and vice versa. This guarantees that every sample, including overlooked anti-hubs, contributes equally to the Uniform Retrieving Matrix $\mathbf{Q}$, making anti-hubs as likely to be retrieved as other normal samples. Based on this equality, the proposed {Uniformity Regularization} loss is formulated as:
\begin{equation}
  \begin{aligned}
    \mathcal{L}_{{Opt}} = -\frac{1}{n} \sum_{i=1}^{n} \sum_{j=1}^{m} \mathbf{Q}_{i,j} \log P(j|i),
  \end{aligned}
  \label{eq:loss_opt}
\end{equation}
where 
\begin{equation}
  \begin{aligned}
P(j|i) = \frac{\exp(\textrm{S}(x_i,y_j))}{\sum_{k=1}^{m} \exp(\textrm{S}(x_i,y_k))}.
  \end{aligned}
  \label{eq:pji}
\end{equation}
And the gradient of $\mathcal{L}_{{Opt}}$ with respect to the similarity scores $\mathbf{S}_{i,j}$ is computed as:
\begin{equation}
  \frac{\partial \mathcal{L}_{\text{Opt}}}{\partial \mathbf{S}_{i,j}} = -\frac{1}{n} \left( \mathbf{Q}_{i,j} - P(j|i) \right).
  \label{eq:gradient_uri}
\end{equation}
Here, $\mathbf{Q}$ is obtained by solving Eq.~(\ref{eq:fair_matching}).
By updating $\mathbf{S}_{i,j}$ to minimize $\mathcal{L}_{\text{Opt}}$, $P(j|i)$ aligns with $\mathbf{Q}_{i,j}$, leading to uniform probabilities $P(j|\cdot) \to \frac{1}{m}$. More details and proof of the losses are provided in the supplementary material.
\begin{figure}[tbp]
\centering
\includegraphics[width=1.00\linewidth]{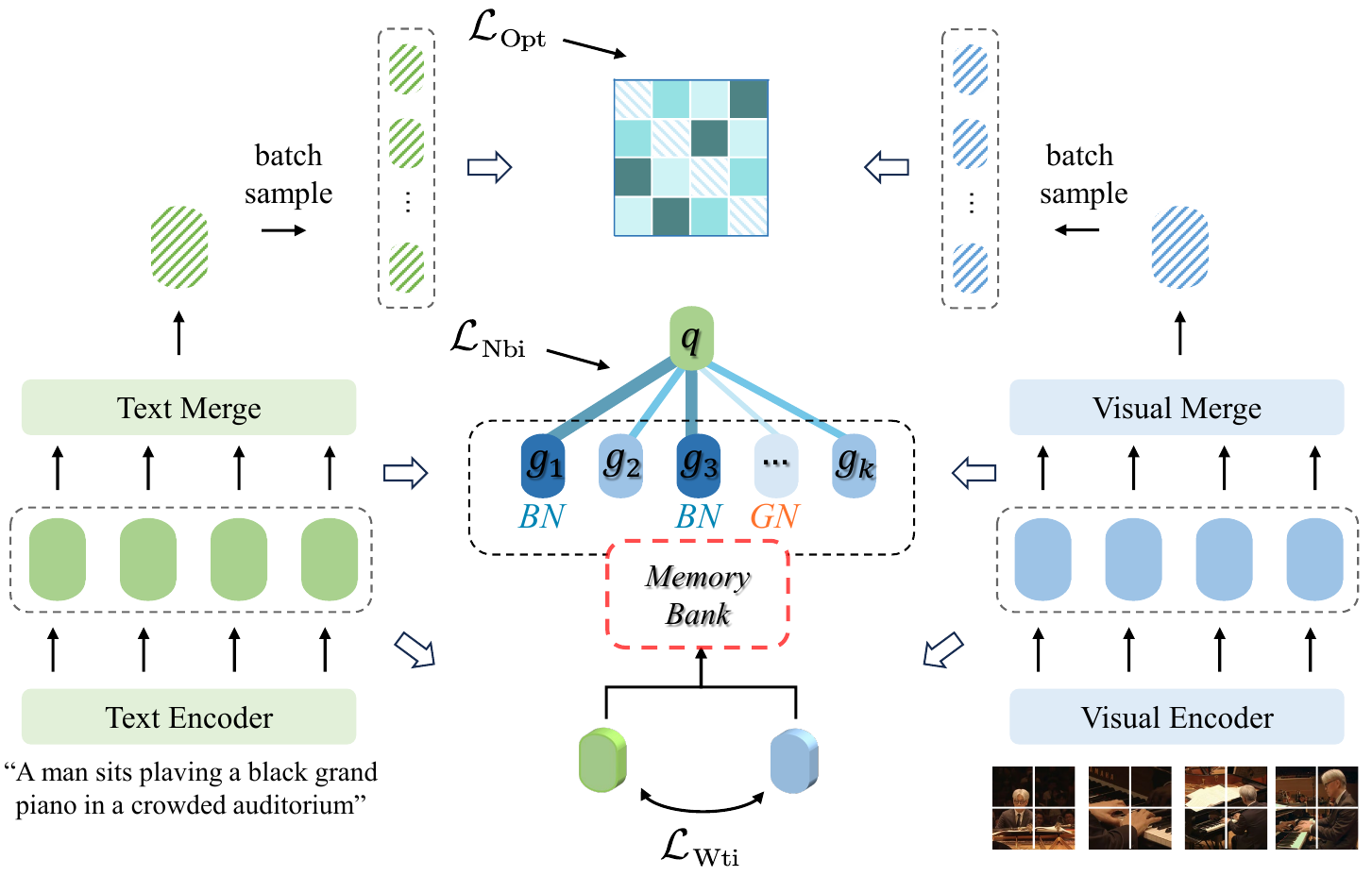}
\caption{Overview of NeighborRetr on two-level hierarchy framework. At the low level, $\mathcal{L}_{\mathrm{Wti}}$ emphasizes the centrality of hubs and $\mathcal{L}_{\mathrm{Nbi}}$ balances neighborhood relations via comparing a memory bank. After token merging, $\mathcal{L}_{\mathrm{Opt}}$ enforces equal retrieval probabilities for overall matching at the high level.}
\label{fig:arch}
\end{figure}

\vspace{0.5em} 
\subsection{Visual-Text Learning}
The framework of NerighborRetr is similar to fine-grained matching methods~\cite{jin2023video,fang2023uatvr}, as shown in Fig.~\ref{fig:arch}. To achieve fine-grained semantic alignment, the visual input $v$ is embedded into patch sequence $\bm{V_p} = \{v^{i}_{p}\}^{N_v}_{i=1}$, where $N_{v}$ is the length of patch $v$. The textual input $t$ is embedded into word sequence $\bm{T_w}=\{t^{j}_{w}\}^{N_t}_{j=1}$, where $N_{t}$ is the length of text $t$. The alignment matrix is defined as: $A=[a_{ij}]^{N_v \times N_t}$, where $a_{ij}=\frac{{v^i_p}^\top t^j_w}{||v^i_p||\ ||t^j_w||}$ represents the alignment score between the $i$-th visual patch and the $j$-th text word. For the $i$-th visual patch, we calculate its maximum alignment score as $\text{max}_{j}a_{ij}$. Then, we use the weighted average maximum alignment score overall visual patch as the visual-to-text similarity using the Weighted Token-wise Interaction (WTI) module~\cite{wang2022disentangled}. Similarly, we can obtain the text-to-visual similarity. The total similarity score can be defined as:
\begin{equation}
\textrm{S}_{\bm{v},\bm{t}}=\frac{1}{2}(\underbrace{\sum_{i=1}^{N_v} \omega_v^i\ \underset{j}{\textrm{max}}\ a_{ij}}_{\textrm{visual-to-text}}+\underbrace{\sum_{j=1}^{N_t} \omega_t^j\ \underset{i}{\textrm{max}}\ a_{ij}}_{\textrm{text-to-visual}}),
\end{equation}
where $[\omega_v^1,...,\omega_v^{N_v}]=\textrm{Softmax}(\textrm{MLP}_v(\bm{V_p}))$ are the weights of the visual patches. We can similarly obtain the text weight. 
To obtain the high-level representation as a single token, we utilize DPC-KNN~\cite{du2016study}, a k-nearest neighbor-based density peaks clustering algorithm, to cluster the visual (textual) tokens for token merging.

\newcommand{\pub}[1]{\color{gray}{\tiny{#1}}}
\newcommand{\Frst}[1]{{\textbf{#1}}}
\newcommand{\Scnd}[1]{{{#1}}}
\newcommand{\Mul}[1]{\color{gray}{#1}}

\begin{table*}[t]
\footnotesize
\centering
\resizebox{1.\linewidth}{!}{
\begin{tabular}{l|cccccc|cccccc}
\toprule[1.25pt]
\multirow{2}{*}{\textbf{Methods}} &\multicolumn{6}{c|}{\textbf{Text-to-Video}} &\multicolumn{6}{c}{\textbf{Video-to-Text}}\\
\cmidrule(rl){2-7}\cmidrule(rl){8-13}
  & R@1$\uparrow$ & R@5$\uparrow$ & R@10$\uparrow$ & Rsum$\uparrow$ & MdR$\downarrow$ & MnR$\downarrow$ & R@1$\uparrow$ & R@5$\uparrow$ & R@10$\uparrow$ & Rsum$\uparrow$ & MdR$\downarrow$ & MnR$\downarrow$ \\ \midrule
T2VLAD~{\cite{wang2021t2vlad}}~\pub{CVPR21} & {29.5} & 59.0 & 70.1 & 158.6 & 4.0 & -  & 31.8 &60.0 &71.1 & 162.9 &\Scnd{3.0} & -\\
CLIP4Clip~{\cite{luo2021clip4clip}}~\pub{Neurocomputing22} & {44.5} & 71.4 & 81.6 & 197.5 &\Frst{2.0} & 15.3 & 42.7 & 70.9& 80.6 & 194.2 & \Frst{2.0} & 11.6\\
X-Pool~{\cite{gorti2022x}}~\pub{CVPR22}  & {46.9} & 72.8 & 82.2 & 201.9 &\Frst{2.0} & 14.3 & 44.4 & 73.3 & 84.0 & 201.7 & \Frst{2.0} & 9.0 \\
TS2-Net~{\cite{liu2022ts2}}~\pub{ECCV22}  & {47.0} & \Scnd{74.5} & {83.8} &{205.3} & \Frst{2.0} & \Scnd{13.0} & 45.3 & 74.1 & 83.7 & 203.1 & \Frst{2.0} & 9.2 \\
EMCL-Net~{\cite{jin2022expectation}}~\pub{NeurIPS22}  & {46.8} & 73.1 &\Scnd{83.1} & 203.0 &\Frst{2.0} & 12.8 & 
46.5 & 73.5 & 83.5 & 203.5 & \Frst{2.0} & 8.8 \\
{DiCoSA~{\cite{jin2023text}}}~\pub{IJCAI23} & \Mul{47.5} & \Mul{74.7} & \Mul{83.8} & \Mul{206.0} & \Mul{2.0} & \Mul{13.2} & \Mul{46.7} & \Mul{75.2} & \Mul{84.3} & \Mul{206.2} & \Mul{2.0} & \Mul{8.9} \\
HBI~\cite{jin2023video}~\pub{CVPR23} & 48.6 & 74.6 & 83.4 & 206.6 & \textbf{2.0} & \textbf{12.0} & 46.8 & \textbf{74.3} & 84.3 & 205.4 & \textbf{2.0} & 8.9 \\
Diffusion~\cite{jin2023diffusionret}~\pub{ICCV23} & 49.0 & \textbf{75.2} & 82.7 & 206.9 & \textbf{2.0} & 12.1 & 47.7 & 73.8 & 84.5 & 206.0 & \textbf{2.0} & 8.8\\
EERCF~\cite{tian2024towards}~\pub{AAAI24} & 47.8 & 74.1 & \textbf{84.1} & 206.0 & -  & - & 44.7 & 74.2 & 83.9 & 202.8 & - & - \\
MPT~\cite{zhang2024mpt}~\pub{ACM MM24} & 48.3 & 72.0 & 81.7 & 202.0 & -  & 14.9 & 46.5 & 74.1 & 82.6 & 203.2 & - & 11.8 \\
\midrule
\rowcolor{aliceblue!60} \textbf{NeighborRetr (Ours)} & \textbf{49.5} & {74.1} & \textbf{84.1} & \textbf{207.7} & \textbf{2.0} &{12.8} & \textbf{48.7} & 74.2 &\Frst{84.7} & \textbf{207.5} &\textbf{2.0} &\textbf{8.4} \\
\bottomrule[1.25pt]
\end{tabular}
}
\caption{Comparisons to state-of-the-art methods on the MSR-VTT dataset. ``$\uparrow$'' denotes higher is better. ``$\downarrow$'' denotes lower is better. DiCoSA~\cite{jin2023text} utilizes QB-Norm~\cite{bogolin2022cross} for inference and is grayed out for a fair comparison.}
\label{tab:comp_msrvtt}
\end{table*}

\begin{table*}[t]
\centering
\small
\setlength{\tabcolsep}{0.0pt}
{
\begin{tabular}{lccccc}
\toprule[1.25pt]
\multicolumn{6}{c}{MSVD}\\
\cmidrule(rl){0-5}
Methods  & R@1$\uparrow$ & R@5$\uparrow$ & R@10$\uparrow$ & MdR$\downarrow$ & MnR$\downarrow$ \\ \midrule
FROZEN & 33.7 & 64.7 & 76.3 & 3.0 &-\\
EMCL-Net & 42.1 & 71.3 & 81.1 & \textbf{2.0} & 17.6\\
CLIP4Clip & 45.2 & 75.5 & 84.3 & \textbf{2.0}  & 10.3 \\
UATVR & 46.0  & 76.3 & 85.1 & \textbf{2.0} & 10.4\\
Diffusion & 46.6 & {75.9} & {84.1} & \textbf{2.0} & 15.7 \\
\midrule
\rowcolor{aliceblue!60} \textbf{Ours} & \textbf{47.9} & \textbf{77.3} & \textbf{86.0} & \textbf{2.0} & \textbf{9.2} \\
\bottomrule[1.25pt]
\end{tabular}
\ 
\begin{tabular}{lccccc}
\toprule[1.25pt]
\multicolumn{6}{c}{ActivityNet Captions}\\
\cmidrule(rl){0-5}
Methods  & R@1$\uparrow$ & R@5$\uparrow$ & R@10$\uparrow$ & MdR$\downarrow$ & MnR$\downarrow$ \\ \midrule
CLIP4Clip & 40.5 & 72.4 & 83.6 & \textbf{2.0}  & 7.5 \\
TS2-Net & 41.0 & 73.6 & 84.5 & \textbf{2.0} & 8.4 \\
{DiCoSA} & \Mul{42.1} & \Mul{73.6} & \Mul{84.6} & \Mul{2.0} & \Mul{6.8} \\
MPT & 41.4 & 70.9 & 82.9 & - & 7.8\\
HBI & 42.2 & 73.0 & 84.6 & \textbf{2.0} & 6.6 \\
\midrule
\rowcolor{aliceblue!60} \textbf{Ours} & \textbf{46.0} &  \Frst{75.9} & \Frst{86.5} & \Frst{2.0} & \Frst{6.1} \\
\bottomrule[1.25pt]
\end{tabular}
\ 
\begin{tabular}{lccccc}
\toprule[1.25pt]
\multicolumn{6}{c}{DiDeMo}\\
\cmidrule(rl){0-5}
Methods  & R@1$\uparrow$ & R@5$\uparrow$ & R@10$\uparrow$ & MdR$\downarrow$ & MnR$\downarrow$ \\ \midrule
FROZEN & 34.6 & 65.0 & 74.7 & 3.0 & - \\
TS2-Net & 41.8 & 71.6 & 82.0 & \textbf{2.0} & 14.8 \\
CLIP4Clip & 42.8 & 68.5 & 79.2 & \textbf{2.0}  & 18.9 \\
Diffusion & 46.7 & 74.7 & 82.7 & \textbf{2.0} & 14.3\\
HBI & 46.9 & 74.9 & 82.7 & \textbf{2.0} & 12.1\\
\midrule
\rowcolor{aliceblue!60} \textbf{Ours} & \textbf{48.2}  & \textbf{76.7} & \textbf{84.9}  & \textbf{2.0} & \textbf{11.9}  \\
\bottomrule[1.25pt]
\end{tabular}}
\caption{Comparisons to other methods on the Text-to-Video task on the MSVD, ActivityNet Captions, and DiDeMo datasets.}
\label{tab:comp_three}
\vspace{-1.0em}
\end{table*}

To stabilize the optimization, following~\cite{jin2023video}, we introduce the KL divergence loss between the distribution of low level $\textrm{S}^{l}_{v,t}$ and high level $\textrm{S}^{g}_{v,t}$, which can be formulated as:
\vspace{-0.5em}
\begin{align}
    \mathcal{L}_{KL}=\text{KL}(\textrm{S}^{g}_{\bm{v},\bm{t}}||\textrm{S}^{l}_{\bm{v},\bm{t}}).
\end{align}
Therefore, the total training objective can be defined as:
\begin{align}
    \mathcal{L}=\frac{1}{2}\sum_{m\in\text{t2v,v2t}} \{
                \mathcal{L}^{m}_{Wti}+ 
                \mathcal{L}^{m}_{Nbi}+
                \mathcal{L}^{m}_{Opt}
                +\mathcal{L}^{m}_{KL}\}.
\end{align}

\section{Experiments}
\subsection{Experimental Settings}
\noindent \textbf{Datasets.}\quad 
\textbf{MSR-VTT}~\cite{xu2016msr} contains 10K videos, each with 20 text descriptions. We follow the training protocol in \cite{gabeur2020multi} and evaluate on the 1K-A testing split. \textbf{ActivityNet Captions}~\cite{krishna2017dense} consists of densely annotated temporal segments of 20K videos. We use the 10K training split and test on the 5K ``val1'' split.  \textbf{DiDeMo}~\cite{anne2017localizing} contains 10K videos with 40K text descriptions. We follow the training and evaluation protocol in \cite{luo2021clip4clip}. \textbf{MSVD}~\cite{chen2011collecting} contains 1,970 videos, we split the train, validation, and test set with 1,200, 100, and 670 videos.
\textbf{ECCV Caption}~\cite{chun2022eccv} is a machine-and-human-verified test, containing $\times 8.5$ positive images and $\times 3.6$ positive captions compared to MS-COCO.
Experiments on \textbf{MS-COCO}~\cite{lin2014microsoft} and \textbf{Flickr30K}~\cite{young2014image} are listed in the supplementary material.

\noindent \textbf{Retrieval Metrics.}\quad
We choose Recall at rank K (R@K), Median Rank (MdR), Sum of Recall at rank \{1, 5, 10\} (Rsum), and mean rank (MnR) to evaluate the retrieval performance. 
We utilize mAP@R~\cite{musgrave2020metric} and R-Precision (R-P)~\cite{musgrave2020metric} metrics on the ECCV Caption to evaluate the ability of the model to recall incorrect negatives.

\noindent \textbf{Hubness Metrics.}\quad
To assess the hubness and the effectiveness of our mitigation strategies, we utilize two kinds of hubness metrics: \textbf{i) Distribution-based Metrics:} K-Skewness (skew)~\cite{radovanovic2010hubs} and K-Skewness Truncated Normal (trunc)~\cite{tomavsev2014role}. Atkinson Index (atkinson)~\cite{fischer2020unequal} and Robin Hood Index (robin)~\cite{feldbauer2018fast}. \textbf{ii) Occurrence-based Metrics:} Antihub Occurrence (anti)~\cite{radovanovic2014reverse} and Hub Occurrence (hub)~\cite{radovanovic2010hubs}. More details on the definition of hubness metrics are in the supplementary material.

\noindent \textbf{Implementation Details.}\quad\label{sec:implementation}
Following previous works~\cite{luo2021clip4clip,jin2022expectation}, we initial the text encoder and visual encoder with
CLIP (ViT-B/32)~\cite{radford2021learning}. The dimensions of the visual and textual representation feature are 512. We use the Adam optimizer~\cite{kingma2014adam} and set the batch size to 128.
The initial learning rate is 1e-4 for MSR-VTT, MSVD, Didemo, and MS-COCO, and 3e-4 for ActivityNet Captions. 
For MSR-VTT and MSVD, the word length is 24 and the frame length is 12, and the network is optimized for 5 epochs.
For ActivityNet Captions and DiDeMo, the word length is 64 and the frame length is 64, and the network is optimized for 10 epochs.
For image datasets, i.e., MS-COCO, the network is optimized for 10 epochs.

\subsection{Comparison with other Methods}
\noindent \textbf{Compared Methods.}\quad
We compare the proposed NeighborRetr with two kinds of text-video retrieval methods: \textbf{i) non-CLIP methods}: T2VLAD~{\cite{wang2021t2vlad}}, FROZEN~{\cite{bain2021frozen}}; and \textbf{ii) CLIP-based methods}: 
CLIP4Clip~{\cite{luo2021clip4clip}}, 
X-Pool~{\cite{gorti2022x}}, 
TS2-Net~{\cite{liu2022ts2}}, 
EMCL-Net~{\cite{jin2022expectation}}, UATVR~\cite{fang2023uatvr}, 
DiCoSA~{\cite{jin2023text}}, 
HBI~\cite{jin2023video},
MPT~\cite{zhang2024mpt}.
We also compare NeighborRetr with text-image retrieval methods: ViLT~\cite{kim2021vilt},
PVSE~\cite{song2019polysemous},
PCME~\cite{chun2021probabilistic},
SGRAF~\cite{diao2021similarity},
CUSA~\cite{huang2024cross},
PCME++~\cite{chun2023improved}.

\begin{table}[t]
\centering
\small
\setlength{\tabcolsep}{1mm}{
\begin{tabular}{l|ccc|ccc}
\toprule[1.25pt]
\multirow{2}{*}{{Methods}} &\multicolumn{3}{c|}{{Text-to-Image}} &\multicolumn{3}{c}{{Image-to-Text}}\\
\cmidrule(rl){2-4}\cmidrule(rl){5-7}
  & R@1$\uparrow$ & R-P$\uparrow$ & mAP@R$\uparrow$ & R@1$\uparrow$ & R-P$\uparrow$ & mAP@R$\uparrow$ \\ \midrule
ViLT & 82.7 & 50.0 & 41.7 & 73.4 & 38.6 & 27.5\\
PVSE & 83.9 & 53.4 & 44.6 & 62.6 & 35.6 & 23.4\\
PCME & 84.1 & 56.3 & 47.9 & 65.5 & 38.7 & 26.2\\
SGRAF & 85.7 & 53.1 & 44.6 & 71.1 & 39.3 & 27.4\\
CUSA & 88.2 & 55.8 & 47.6 & 80.9 & 44.1 & 33.6\\
PCME++ & 90.8 & 57.0 & 49.5 & 81.3 & 45.0 & \textbf{34.4}\\
\midrule
\rowcolor{aliceblue!60} \textbf{Ours} & \textbf{92.1} & \textbf{57.7} & \textbf{49.8} & \textbf{82.1} & \textbf{45.6} & \textbf{34.4} \\
\bottomrule[1.25pt]
\end{tabular}}
\caption{Comparisons to others on the ECCV Caption datasets.}
\label{tab:comp_eccv_caption}
\vspace{-1em}
\end{table}

\begin{table}[t]
\centering
\small
\setlength{\tabcolsep}{1mm}{
\begin{tabular}{lcccccccc}
\toprule[1.25pt]
Methods & skew$\downarrow$ & trunc$\downarrow$ & atkinson$\downarrow$ & robin$\downarrow$ & anti$\downarrow$ & hub$\downarrow$ \\ \midrule
X-Pool & 5.14 & 1.24 & 0.71 & 0.70 & 0.51 & 0.80\\
EMCL-Net & 4.67 & 1.23 & 0.71 & 0.70 & 0.53 & 0.78\\
DiCoSA & 7.37 & 1.23 & 0.59 & 0.64 & 0.34 & 0.71\\
HBI & 5.19 & 1.26 & 0.75 & 0.75 & 0.55 & 0.84\\
DiffusionRet & 5.20 & 1.26 & 0.75 & 0.75 & 0.55 & 0.84\\
UCOFIA & 7.10 & 1.40 & 0.92 & 0.90 & 0.78 & 0.94\\
\midrule
Baseline\dag & 5.97 & 1.30 & 0.81 & 0.79 & 0.63 & 0.80\\
\rowcolor{aliceblue!60} \textbf{Ours} & \textbf{3.20} & \textbf{1.04} & \textbf{0.44} & \textbf{0.45} & \textbf{0.23} & \textbf{0.58} \\
\bottomrule[1.25pt]
\end{tabular}}
\caption{Comparing with other methods on the Text-to-Video task on MSR-VTT for hubness metrics. \dag the same two-level hierarchy model with only a vanilla contrastive loss and a KL loss.}
\label{comp_hub_others}
\vspace{-1.0em}
\end{table}


\noindent \textbf{Retrieval Performance.}\quad
In Table.~\ref{tab:comp_msrvtt}, we compare NeighborRetr with existing methods on the MSR-VTT dataset. NeighborRetr achieves the best performance across most metrics in both text-to-video (T2V) and video-to-text (V2T) retrieval tasks.
Compared to the generative diffusion-based method DiffusionRet~\cite{jin2023diffusionret}, NeighborRetr shows a clear improvement of $1.0\%$ on V2T R@1 metric. When compared to the fine-grained interaction method HBI~\cite{jin2023video}, which employs a three-level framework, NeighborRetr with only a two-level framework achieves $0.9\%$ improvement on T2V R@1 and $1.9\%$ improvement on V2T R@1.
Table.~\ref{tab:comp_three} compares NeighborRetr with other methods on the text-to-video retrieval task on MSVD, ActivityNet Captions, and DiDeMo datasets. 
On the short video dataset MSVD, NeighborRetr surpasses other CLIP-based methods by $1.3\%$. 
On long video datasets ActivityNet and DiDeMo, NeighborRetr consistently outperforms others, advancing the R@1 by $3.8\%$ and $1.3\%$. 
Additionally, as shown in Table.~\ref{tab:comp_eccv_caption}, NeighborRetr surpasses the previous method by $1.3\%$ in R@1 for text-to-image retrieval and by $0.8\%$ in image-to-text retrieval on the ECCV Caption dataset.
More retrieval results are listed in the supplementary material.

\noindent\textbf{Hubness Mitigation.}\quad
Alongside its strong retrieval performance, NeighborRetr significantly mitigates the hubness problem as evidenced in Table~\ref{comp_hub_others}. NeighborRetr reduces skew by $46\%$ compared to the baseline, 
effectively shortening the long tail of the distribution of retrieval frequency.
The $43\%$ decrease in robin indicates a more balanced and equitable distribution. Additionally, the $28\%$ reduction of hub demonstrates NeighborRetr's ability to prevent over-centralized samples, enhancing retrieval fairness.

\setlength{\abovecaptionskip}{0.8em} 
\begin{figure}[t]
\centering
\includegraphics[width=1.\linewidth]{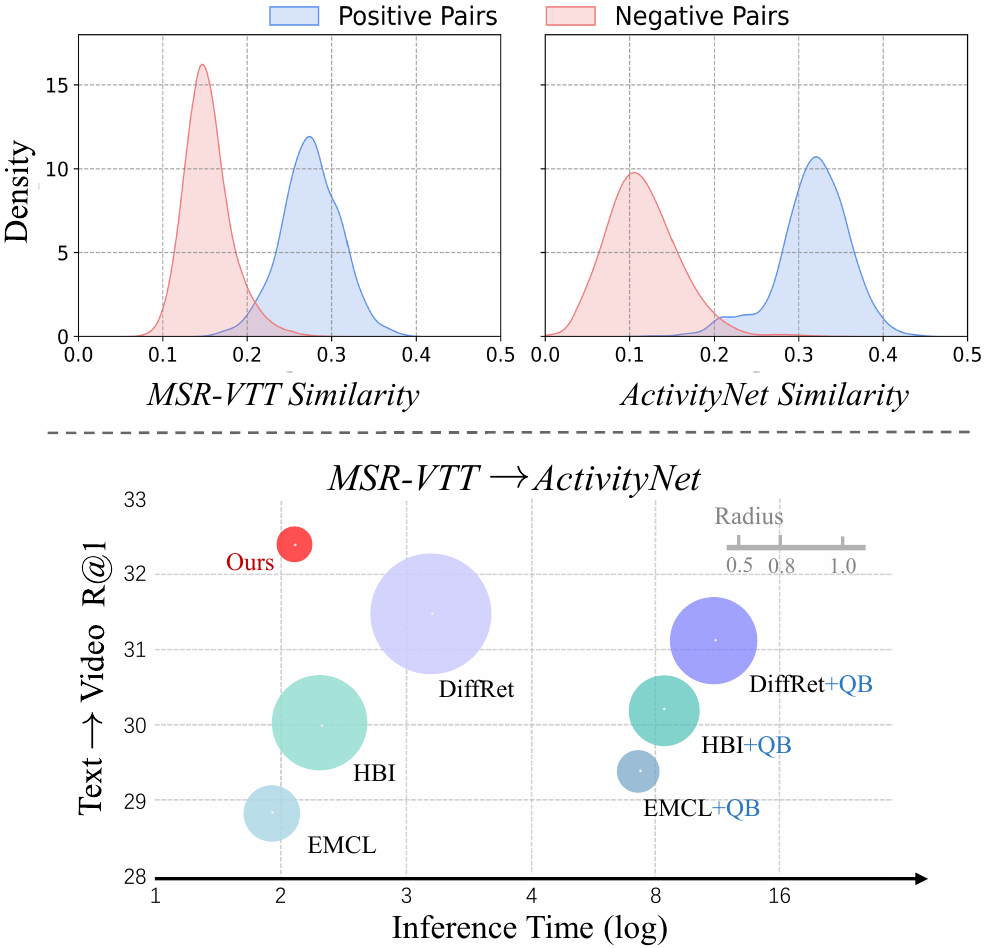}
\vspace{-1em}
\caption{\emph{Top:} Distributions of cross-modal similarity of MSR-VTT and ActivityNet. \emph{Bottom:} Cross-domain adaptation on MSRVTT$\rightarrow$ActivityNet. The radius denotes hub occurrence~\cite{radovanovic2010hubs}.}
\label{fig:qipao}
\vspace{-0.5em}
\end{figure}

\noindent\textbf{Cross-domain Adaptation.}\quad
At the top of Fig.~\ref{fig:qipao}, we show the substantial domain gap between MSRVTT (source) and ActivityNet (target), urging methods for robust cross-domain generalization. To evaluate this, we train NerighborRetr on the MSR-VTT and test it on the ActivityNet test set.
As shown at the bottom of Fig.~\ref{fig:qipao}, our method achieves the lowest hub occurrence and the best retrieval performance on MSR-VTT $\rightarrow$ ActivityNet, indicating that addressing hubness during training benefits cross-domain adaptation. Moreover, we compare with QB-Norm~\cite{bogolin2022cross}, a test-time adjusting method that recalibrates retrieval scores by referencing a query bank from the training set. Due to the large distribution shift, QB-Norm can not effectively use training set priors to alleviate the hubness problem in the test domain, limiting its retrieval performance.


\begin{table}[t]
\centering
\footnotesize 
\setlength{\tabcolsep}{0.3mm} 
{
\begin{tabular}{ccccccccccc}
\toprule[1.25pt]
$\mathcal{L}_{Wti}$ & $\mathcal{L}_{Opt}$ & $\mathcal{L}_{Nbi}$ & $\mathcal{L}_{KL}$ & R@1$\uparrow$ & R@5$\uparrow$ & R@10$\uparrow$ & Rsum$\uparrow$ & MnR$\downarrow$ & anti$\downarrow$ & hub$\downarrow$ \\ \midrule
\Checkmark & \ding{55} & \ding{55} & \ding{55} & 47.2 & 74.0 & 83.5 & 204.7 & 12.9 & 0.63 & 0.83 \\
\Checkmark & \Checkmark & \ding{55} & \ding{55} & 48.0 & 73.8 & 83.7 & 205.5 & 12.5 & 0.32 & 0.76 \\
\Checkmark & \ding{55} & \Checkmark & \ding{55} & 48.3 & 74.5 & 83.9 & 206.7 & 12.5 & 0.45 & 0.62 \\
\Checkmark & \Checkmark & \ding{55} & \Checkmark & 48.4 & 73.9 & 83.7 & 206.0 & \textbf{12.3} & 0.28 & 0.74 \\
\rowcolor{aliceblue!60}\Checkmark & \Checkmark & \Checkmark & \Checkmark & \textbf{49.5} & \textbf{74.1} & \textbf{84.1} & \textbf{207.7} & 12.8 & \textbf{0.23} & \textbf{0.52} \\ 
\bottomrule[1.25pt]
\end{tabular}}
\caption{Retrieval performance and hubness metric on the Text-to-Video task on MSR-VTT for all losses ablation.}
\label{tab:loss_alb}
\vspace{-1.5em}
\end{table}

\begin{table}[t]
\centering
\small
\setlength{\tabcolsep}{0.4mm}{
\begin{tabular}{cccccccc}
\toprule[1.25pt]
$C_{\mathcal{L}_{Wti}}$ & $C_{\mathcal{L}_{Nbi}}$ & R@1$\uparrow$ & R@5$\uparrow$ & R@10$\uparrow$ & Rsum$\uparrow$ & MdR$\downarrow$ & MnR$\downarrow$ \\ \midrule
intra & intra & 48.3 & 73.7 & 83.4 & 205.4 & 2.0 & \textbf{12.2}\\
inter & inter & 47.9 & {73.8} & {83.7} & 205.4 & 2.0 & 13.0\\
inter & intra & 47.7 & 73.2 & 83.2 & 204.1 & 2.0 & {12.5}\\
\rowcolor{aliceblue!60} \textbf{intra} & \textbf{inter} & \textbf{49.5} & \textbf{74.1} & \textbf{84.1} & \textbf{207.7} & \textbf{2.0} & {12.8} \\
\bottomrule[1.25pt]
\end{tabular}}
\caption{Retrieval performance on the Text-to-Video task on MSR-VTT for diverse centrality.}
\label{tab:connect_intra_inter}
\vspace{-0.5em}
\end{table}

\subsection{Ablation Studies}
\noindent\textbf{Contribution of each loss of NeighborRetr.}\quad 
In Table~\ref{tab:loss_alb}, we compare different combinations of loss functions: $\mathcal{L}_{Wti}$, $\mathcal{L}_{Opt}$, $\mathcal{L}_{Nbi}$, and $\mathcal{L}_{KL}$. 
The ablation reveals that incorporating $\mathcal{L}_{Wti}$ (Row~1) improves Rsum to 204.7, boosting discriminative power by centralizing samples.
Adding $\mathcal{L}_{Opt}$ (Row~2) increases Rsum by $0.8\%$, with \textit{anti} decreasing by 49\% and \textit{hub} decreasing by 8\% compared to $\mathcal{L}_{Wti}$ (Row~1), which in particular affects the uniform probability of anti-hub.
The $\mathcal{L}_{Nbi}$ loss increases Rsum by $2.0\%$, with \textit{anti} decreasing by 29\% and \textit{hub} decreasing by 25\% when paired with $\mathcal{L}_{Wti}$ (Row~3), balancing good and bad hubs in neighborhood relations.
Utilizing all these losses (Row~5) yields the best results, showcasing their complementary strengths: $\mathcal{L}_{Wti}$ enhances central samples, $\mathcal{L}_{Opt}$ realigns retrieval distribution, and $\mathcal{L}_{Nbi}$ balances neighborhood relations.

\noindent\textbf{Intra/Inter-Modal Centrality for $\mathcal{L}_{Wti}$ and $\mathcal{L}_{Nbi}$.}\quad
To validate the effectiveness of our different measurements of centrality for $\mathcal{L}_{Wti}$ and $\mathcal{L}_{Nbi}$, we compare our methods trained with different choices of centralities $C_{\mathcal{L}_{Wti}}$ and $C_{\mathcal{L}_{Nbi}}$. In Table.~\ref{tab:connect_intra_inter}, we opt for measuring intra/inter-modal centrality ($C$ or $\hat{C}$) in $\mathcal{L}_{Wti}$ and $\mathcal{L}_{Nbi}$. 
Results demonstrate that employing intra-modal centrality for $\mathcal{L}_{Wti}$ and inter-modal centrality for $\mathcal{L}_{Nbi}$ achieves the best performance compared to other measurement methods, with the final Rsum addition improvement by $3.6\%$.
Integrating intra-modal centrality into $\mathcal{L}_{Wti}$ preserves each modality’s structure, by promoting central samples as hubs, thereby capturing intrinsic representations.
Meanwhile, applying inter-modal centrality to $\mathcal{L}_{Nbi}$ leverages cross-modal embedding density, improving the discrimination of neighborhoods based on inter-modal centrality.

\setlength{\abovecaptionskip}{0.5em} 
\begin{figure}[t]
\centering
\includegraphics[width=1.\linewidth]{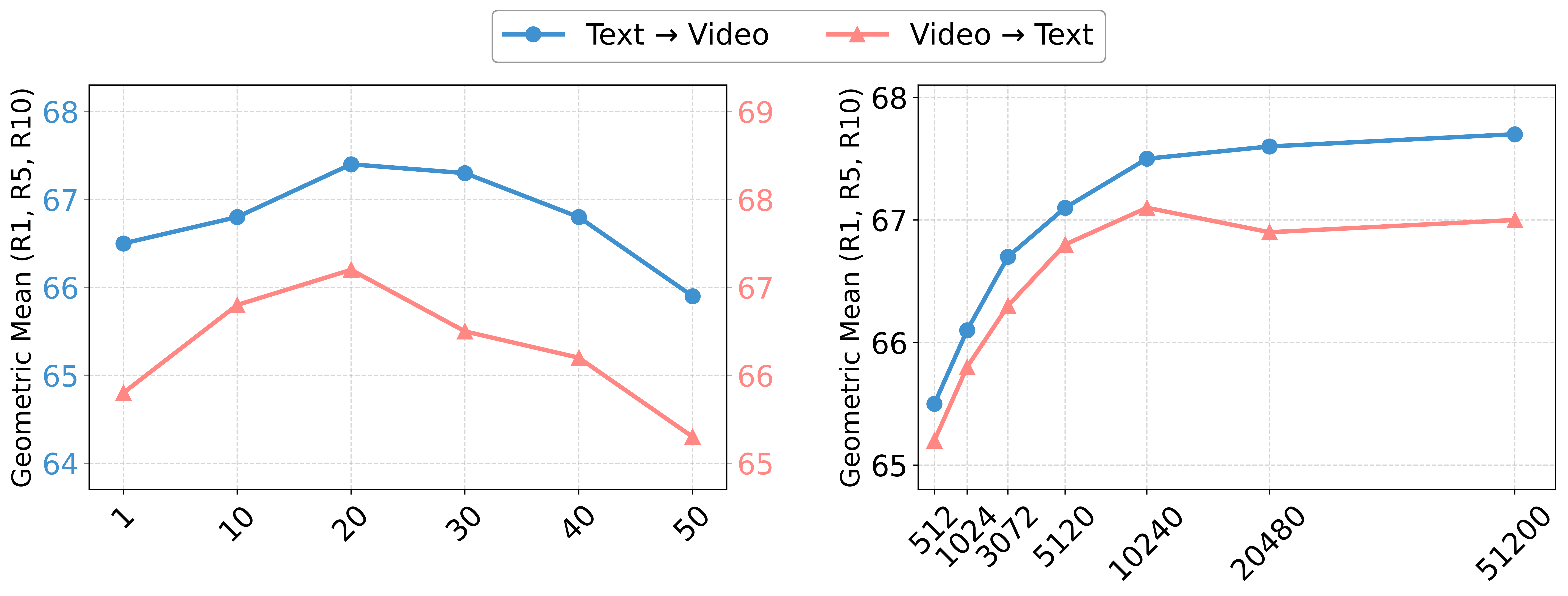}
\caption{Retrieval performance on MSR-VTT: The y-axis shows the geometric mean of R@\{1,5,10\}. \textit{Left}: Ablation study on the size of neighbors; the x-axis represents the size of neighbors. \textit{Right}: Ablation study on the size of the Memory Bank during training; the x-axis represents the Memory Bank size.}
\label{fig:nbi_mem_ablations}
\vspace{-0.5em}
\end{figure}

\noindent\textbf{Different Formulations of $\mathcal{L}_{Nbi}$.}\quad
To evaluate our neighborhood loss design (Simi-Cent), we ablate various formulations in Table~\ref{tab:nighbor_loss_abl}. 
We compared the individual effects of cross-modal similarity (Simi) and centrality (Cent) scores, as well as their additive combination (Simi+Cent), against a baseline model without $\mathcal{L}_{Nbi}$. The results in Rows 2\&3 show that using Simi or Cent alone provides minimal improvement over the baseline, indicating that adjusting rankings by either similarity or centrality alone is insufficient.
However, the additive combination Simi+Cent yields only a slight boost in retrieval performance (Row 4), underscoring the limited effectiveness of this approach compared to subtraction. 
Our proposed Simi-Cent formulation, which subtracts centrality from similarity, achieves the best results (Row 5). This formulation effectively brings beneficial neighbors closer while distancing detrimental hubs, enhancing the embedding space and improving retrieval accuracy.

\begin{table}[t]
\centering
\small
\setlength{\tabcolsep}{0.6mm}{
\begin{tabular}{ccccccc}
\toprule[1.25pt]
Methods  & R@1$\uparrow$ & R@5$\uparrow$ & R@10$\uparrow$ & Rsum$\uparrow$ & MdR$\downarrow$ & MnR$\downarrow$ \\ \midrule
w/o $\mathcal{L}_{Nbi}$ & 48.4 & 73.9 & 83.7 & 206.0 & \textbf{2.0} & 12.3\\
Simi& 48.6 & 74.1 & 83.8 & 206.5 & \textbf{2.0} & \textbf{12.2}\\
Cent & 48.2 & 74.0 & 83.8 & 206.0 & \textbf{2.0} & 12.4\\
Simi+Cent & 48.0 & 73.4 & 83.6 & 205.0 & \textbf{2.0} & 12.8\\
\rowcolor{aliceblue!60} \textbf{Simi-Cent} & \textbf{49.5} & \textbf{74.1} & \textbf{84.1} & \textbf{207.7} & \textbf{2.0} & {12.8}\\
\bottomrule[1.25pt]
\end{tabular}}
\caption{Retrieval performance on the Text-to-Video task on MSR-VTT for the neighborhood loss $\mathcal{L}_{Nbi}$ ablation.}
\label{tab:nighbor_loss_abl}
\vspace{-1em}
\end{table}


\noindent\textbf{Size of Neighborhood and Memory Bank.}\quad
As illustrated on the left of Fig.~\ref{fig:nbi_mem_ablations}, increasing the number of neighbors initially enhances retrieval performance, reaching an optimal value at 20 neighbors, beyond which performance declines.
This suggests that incorporating too many neighbors may weaken the regularization due to sparse relevant examples. 
On the right of Fig.~\ref{fig:nbi_mem_ablations}, we observe that performance steadily improves with larger memory bank sizes, gradually saturating beyond 10,240. Consequently, we set the memory bank size to 10,240 in our experiments.



\section{Conclusions}
In this paper, we introduced NeighborRetr, a method to systematically address the hubness problem in cross-modal retrieval by distinguishing between good hubs and bad hubs. By identifying hubs based on sample centrality, we perform centrality weighting and adaptively balance neighborhood relationships using a neighbor adjusting loss to promote good hubs and penalize bad ones. Additionally, we incorporate a uniform regularization loss enforcing uniform marginal constraints to treat anti-hubs equally. Experiments on various benchmarks demonstrate that NeighborRetr effectively mitigates hubness, resisting distribution shifting, and achieves state-of-the-art performance.

\vspace{0.4em}
\noindent\textbf{Acknowledgements}
This research was supported by Natural Science Foundation of China (No. 62302453) and Zhejiang Provincial Natural Science Foundation of China (No. LMS25F020003). Thank reviewer and Peng Jin for advice.
\twocolumn
\bibliographystyle{ieeenat_fullname}
\bibliography{main}

\clearpage



\appendix
{\noindent\LARGE\bfseries Supplementary Material}
\vspace{0.5em}
\section{Method Details}
\subsection{Hubness Observation}
In Sec. 3.2, to categorize unlabeled cross-modal pairs as pseudo-positive or negative, we propose using intra-text similarity as a probe to identify potential cross-modal multi-correlations. Here, we proved further rationale for using the text encoder in the CLIP for probing. In ~\cite{li2023integrating}, to calculate the relevance degree between unlabeled samples, they compare across several ways for false negative identification and find that intra-text similarities are the best way to discourse false negatives from true negatives.  Especially, as shown in Fig.4 of~\cite{li2023integrating}, i) cross-modal comparison (text encoder in CLIP v.s. ViT in CLIP) can judge true negatives and false negatives but the two data distributions do not separate by a large margin. ii) Intra-visual comparison (ViT in CLIP v.s. ViT in CLIP) cannot accurately distinguish true negatives from false negatives. iii) Intra-text comparison (text encoder in CLIP v.s. text encoder in CLIP) has the ability to discriminate false negatives, as text is an abstraction of semantic content and the text similarity is closer to the semantic content similarity. In our work, we find that the distribution of intra-text comparison is well-aligned with the ground truth positive-negative annotations. In a broader context, especially without multi-correlations annotations, intra-text similarity can be used as a probe of semantic alignment.

\subsection{Proof of Gradient}
\noindent \textbf{Gradient of $\mathcal{L}_{\text{Nbi}}$.}
The gradient of the Neighbor Adjusting Loss $\mathcal{L}_{\text{Nbi}}(x_i)$ with respect to the similarity score $\textrm{S}(x_i, y_j)$ is derived as follows:
\vspace{1.0em}
\begin{flalign*}
& \quad \frac{\partial \mathcal{L}_{\text{Nbi}}(x_i)}{\partial \textrm{S}(x_i, y_j)} \\
&= \frac{\partial}{\partial \textrm{S}(x_i, y_j)} \left( - \sum_{y_k \in \mathcal{N}^+(x_i)} \mathcal{H}(y_k) \log P(y_k \mid \mathcal{N}^+(x_i)) \right ) \\
&= - \sum_{y_k \in \mathcal{N}^+(x_i)} \mathcal{H}(y_k) \frac{\partial}{\partial \textrm{S}(x_i, y_j)} \log P(y_k \mid \mathcal{N}^+(x_i)) \\
&= - \sum_{y_k \in \mathcal{N}^+(x_i)} \mathcal{H}(y_k) \left( \delta_{kj} - P(y_j \mid \mathcal{N}^+(x_i)) \right ) \\
&= - \mathcal{H}(y_j) (1 - P(y_j \mid \mathcal{N}^+(x_i))) \\
&\quad + P(y_j \mid \mathcal{N}^+(x_i)) \sum_{y_{k} \in \mathcal{N}^+(x_i)} \mathcal{H}(y_k) \\
&= - \mathcal{H}(y_j) + P(y_j \mid \mathcal{N}^+(x_i)) \sum_{y_{k} \in \mathcal{N}^+(x_i)} \mathcal{H}(y_k) \\
&= P(y_j \mid \mathcal{N}^+(x_i)) - \mathcal{H}(y_j).
\end{flalign*}
\vspace{0.1em}

\noindent The third step involves recognizing that the derivative of $\log P(y_k \mid \mathcal{N}^+(x_i))$ with respect to $\textrm{S}(x_i, y_j)$ is $\delta_{kj} - P(y_j \mid \mathcal{N}^+(x_i))$, where $\delta_{kj}$ is the Kronecker delta function. 
The derivation reveals that the gradient equals $P(y_j \mid \mathcal{N}^+(x_i)) - \mathcal{H}(y_j)$, indicating that it is the difference between the predicted probability and the adjusted target. This guides the update of similarity scores $\textrm{S}(x_i, y_j)$: a positive gradient, occurring when the predicted probability $P(y_j \mid \mathcal{N}^+(x_i)) $ exceeds the adjusted target $\mathcal{H}(y_j)$, decreases the similarity score to penalize bad hubs, while a negative gradient increases it to promote good hubs, thereby balancing the representation space.

\noindent \textbf{Gradient of $\mathcal{L}_{\text{Opt}}$.}
Derive the gradient of the Uniformity Regularization loss $\mathcal{L}_{\text{Opt}}$ with respect to the similarity score $\mathbf{S}_{i,j}$ for query $i$ and gallery $j$. 
\vspace{0.4em}
\begin{flalign*}
& \quad \frac{\partial \mathcal{L}_{\text{Opt}}}{\partial \mathbf{S}_{i,j}} \\
&= \frac{\partial}{\partial \mathbf{S}_{i,j}} \left( -\frac{1}{n} \sum_{i'=1}^{n} \sum_{l=1}^{m} \mathbf{Q}_{i',l} \log P(l|i') \right ) \\
&= -\frac{1}{n} \frac{\partial}{\partial \mathbf{S}_{i,j}} \left( \sum_{i'=1}^{n} \sum_{l=1}^{m} \mathbf{Q}_{i',l} \log P(l|i') \right ) \\
&= -\frac{1}{n} \left( \sum_{l=1}^{m} \mathbf{Q}_{i,l} \frac{\partial}{\partial \mathbf{S}_{i,j}} \log P(l|i) \right ) \\
&= -\frac{1}{n} \left( \sum_{l=1}^{m} \mathbf{Q}_{i,l} \left( \delta_{j,l} - P(j|i) \right ) \right ) \\
&= -\frac{1}{n} \left( \mathbf{Q}_{i,j} - P(j|i) \sum_{l=1}^{m} \mathbf{Q}_{i,l} \right ) \\
&= -\frac{1}{n} \left( \mathbf{Q}_{i,j} - P(j|i) \right ).
\end{flalign*}
\vspace{0.4em}

\noindent The fourth step involves recognizing that the derivative of $\log P(l|i)$ with respect to $\mathbf{S}_{i,j}$ is $\delta_{j,l} - P(j|i)$, where $\delta_{j,l}$ is the Kronecker delta function. 
Since $\mathbf{Q}{i,j}$ is designed to uniform probability retrieval, the gradient encourages the model to adjust $P(j|i)$ towards a uniform distribution over all possible retrievals. By minimizing the loss, the optimization process reduces this discrepancy, effectively guiding the predicted probabilities to match the uniform target. This mechanism ensures that the model will not favor hub or ignore anti-hub.

\begin{table*}[t]
\footnotesize
\centering
\setlength{\tabcolsep}{3pt} 
\begin{minipage}{0.48\linewidth}
\centering
\resizebox{1.\linewidth}{!}{
\begin{tabular}{l|ccc|ccc|c}
\toprule[1.25pt]
\multirow{3}{*}{Model} & \multicolumn{7}{c}{\textbf{Flickr30K (1K Test Set)}}\\
\cmidrule(rl){2-8}
 & \multicolumn{3}{c|}{Text-to-Image} & \multicolumn{3}{c|}{Image-to-Text} & \multirow{2}{*}{Rsum$\uparrow$} \\
 & R@1$\uparrow$ & R@5$\uparrow$ & R@10$\uparrow$ & R@1$\uparrow$ & R@5$\uparrow$ & R@10$\uparrow$ & \\
\midrule
CSIC & 88.5 & 98.1 & 99.4 & 75.3 & 93.6 & 96.7 & 551.6 \\
IAIS & 88.3 & 98.4 & 99.4 & 76.9 & 93.3 & 95.7 & 552.0 \\
ViSTA & 89.5 & 98.4 & 99.6 & 75.8 & 94.2 & 96.9 & 554.4 \\
ViLEM & 92.4 & 99.2 & 99.7 & 78.1 & 94.6 & 97.0 & 561.0 \\
I-CLIP & 91.2 & 99.2 & 99.7 & 74.3 & 92.3 & 95.5 & 552.2 \\
CUSA & 90.8 & 99.1 & 99.7 & 77.4 & 95.5 & 97.7 & 560.2 \\
\midrule
\rowcolor{aliceblue!60} \textbf{Ours} & \textbf{92.7} & \textbf{99.3} & \textbf{99.8} & \textbf{79.5} & \textbf{95.7} & \textbf{97.8} & \textbf{564.8} \\
\bottomrule[1.25pt]
\end{tabular}
}
\end{minipage}\hfill
\begin{minipage}{0.49\linewidth}
\centering
\resizebox{1.\linewidth}{!}{
\begin{tabular}{l|ccc|ccc|c}
\toprule[1.25pt]
\multirow{3}{*}{Model} & \multicolumn{7}{c}{\textbf{MSCOCO (5K Test Set)}}\\
\cmidrule(rl){2-8}
 & \multicolumn{3}{c|}{Text-to-Image} & \multicolumn{3}{c|}{Image-to-Text} & \multirow{2}{*}{Rsum$\uparrow$} \\
 & R@1$\uparrow$ & R@5$\uparrow$ & R@10$\uparrow$ & R@1$\uparrow$ & R@5$\uparrow$ & R@10$\uparrow$ & \\
\midrule
UNITER & 63.3 & 87.0 & 93.1 & 48.4 & 76.7 & 85.9 & 454.4 \\
ViSTA & 63.9 & 87.8 & 93.6 & 47.8 & 75.8 & 84.5 & 453.4 \\
SOHO & 66.4 & 88.2 & 93.8 & 50.6 & 78.0 & 86.7 & 463.7 \\
PCME & 65.3 & 89.2 & 94.5 & 51.2 & 79.1 & 87.5 & 474.2 \\
ViLEM & 69.0 & \textbf{90.7} & 95.1 & 52.6 & 79.4 & 87.2 & 474.0 \\
CUSA & 67.9 & 90.3 & 94.7 & 52.4 & \textbf{79.8} & 88.1 & 473.2 \\
\midrule
\rowcolor{aliceblue!60} \textbf{Ours} & \textbf{69.5} & 90.5 & \textbf{95.3} & \textbf{53.2} & 79.7 & \textbf{88.2} & \textbf{476.4} \\
\bottomrule[1.25pt]
\end{tabular}
}
\end{minipage}
\caption{Experimental results of image-text retrieval on MSCOCO and Flickr30K datasets. ``$\uparrow$'' denotes higher is better.}
\label{tab:itr_combined}
\vspace{-0.5em}
\end{table*}

\subsection{Uniform Regularization}
In Sec. 4.4, we utilize a uniform marginal constraint inspired by Sinkhorn Distances~\cite{cuturi2013sinkhorn} to enforces equal retrieval probabilities for overall matching at the high level.

We start by deriving high-level representations for each modality using the Deep Projection Clustering with K-Nearest Neighbors (DPC-KNN)~\cite{du2016study} module. To learn high semantic alignment, high-level representations are denoted as $\mathbf{V_g} = \{v^{i}_{g}\}_{i=1}^{N}$ for the visual modality and $\mathbf{T_g} = \{t^{j}_{g}\}_{j=1}^{N}$ for the textual modality, where $N$ represent the lengths of high-level representation. The high similarity between a visual high-level representation $v_g^i$ and a textual high-level representation $t_g^j$ is then computed using a cosine similarity metric. The high alignment matrix $G$ is defined as $G=[g_{ij}]^{N \times N}$, where:


\begin{equation}
    g_{ij}=\frac{{v^i_g}^\top t^j_g}{||v^i_g||\ ||t^j_g||},
\end{equation}
which is the high similarity score between the $i$-th visual high representation and the $j$-th textual high representation.


Uniform Retrieving Matrix $\mathbf{Q}^*$ has a simple normalized exponential matrix solution by Sinkhorn fixed point iterations~\cite{cuturi2013sinkhorn}.
By extensively reordering relationships, optimal transport aligns semantically related pairs more closely, balancing the centrality of hubs and anti-hubs. To further enhance robustness, realigned targets $\mathbf{Q}$ are incorporated:

\begin{align}
    \mathbf{Q} = (1-\beta)\mathbf{I}_{B}+\beta \mathbf{Q}^*.
\end{align}
Here, $\mathbf{I}_{B}$ is the identity matrix of size $B$, where $B$ represents the batch size, and $\beta \in [0,1]$ controls the balance between the identity matrix and the uniform retrieving matrix $\mathbf{Q}^*$. By redefining $\mathbf{Q}$ in this manner, each query or gallery is realigned based on uniform regularization, thereby enforcing balanced retrieval probabilities across all samples.

\section{More Experiments}
\subsection{Experimental Settings}
\noindent \textbf{Datasets.}\quad 
We also evaluate our method on two text-image retrieval benchmarks: Flickr30K and MS-COCO. \textbf{Flickr30K}~\cite{young2014image} contains 31K images, each image is annotated with 5 sentences. Following the data split of~\cite{lee2018stacked}, we use 1K images for validation, 1K images for testing, and the remaining 29K for training. \textbf{MS-COCO}~\cite{lin2014microsoft} contains 123K images, and each image comes with 5 sentences. We mirror the data split setting of~\cite{faghri2017vse++}. We use 113K images for training, 5K images for validation, and 5K images for testing.

\noindent \textbf{Implementation Details.}\quad
For Flickr30K and MS-COCO, we utilize the CLIP (ViT-B/16) ~\cite{radford2021learning} as the pre-trained model. The dimensions of the visual and textual representation feature are 512. We use the Adam optimizer~\cite{kingma2014adam} and set the batch size to 128.
The initial learning rate is 1e-4 for Flickr30K, and MS-COCO. 
For Flickr30K, the network is optimized for 5 epochs.
For MS-COCO, the network is optimized for 10 epochs.

\begin{table*}[t]
\centering
\small
\setlength{\tabcolsep}{0.0pt}
{
\begin{tabular}{lccccc}
\toprule[1.25pt]
\multicolumn{6}{c}{MSVD}\\
\cmidrule(rl){0-5}
Methods  & R@1$\uparrow$ & R@5$\uparrow$ & R@10$\uparrow$ & MdR$\downarrow$ & MnR$\downarrow$ \\ \midrule
EMCL-Net & 54.3 & 81.3 & 88.1 & \textbf{1.0} & 5.6\\
CLIP4Clip & 62.0 & 87.3 & 92.6 & \textbf{1.0} & 4.5\\
X-Clip & 60.9 & 87.8 & - & - & 4.7\\
Diffusion & 61.9 & 88.3 & 92.9 & \textbf{1.0} & 4.5\\
\midrule
\rowcolor{aliceblue!60} \textbf{Ours} &  \textbf{63.3} & \textbf{89.6} & \textbf{95.4} & \textbf{1.0} & \textbf{3.3}\\
\bottomrule[1.25pt]
\end{tabular}
\ 
\begin{tabular}{lccccc}
\toprule[1.25pt]
\multicolumn{6}{c}{ActivityNet Captions}\\
\cmidrule(rl){0-5}
Methods  & R@1$\uparrow$ & R@5$\uparrow$ & R@10$\uparrow$ & MdR$\downarrow$ & MnR$\downarrow$ \\ \midrule
CLIP4Clip & 41.4 & 73.7 & 85.3 & \textbf{2.0} & 6.7\\
EMCL-Net & 42.7 & 74.0 & - & \textbf{2.0} &-\\
ECLIPSE & 42.3 & 73.2 & 83.8 & - & 8.2 \\
HBI & 42.2 & 73.0 & 86.0 & \textbf{2.0} & 6.5 \\
\midrule
\rowcolor{aliceblue!60} \textbf{Ours} & \textbf{43.5} & \textbf{74.6} & \textbf{86.7} & \textbf{2.0} & \textbf{6.1} \\
\bottomrule[1.25pt]
\end{tabular}
\ 
\begin{tabular}{lccccc}
\toprule[1.25pt]
\multicolumn{6}{c}{DiDeMo}\\
\cmidrule(rl){0-5}
Methods  & R@1$\uparrow$ & R@5$\uparrow$ & R@10$\uparrow$ & MdR$\downarrow$ & MnR$\downarrow$ \\ \midrule
X-Clip & 43.1 & 77.2 & - & - & 10.9\\
EMCL-Net & 45.7 & 74.3 & 82.7 & \textbf{2.0} & 10.9 \\
Diffusion & 46.2 & 74.3 & 82.2 & \textbf{2.0} & 10.7\\
HBI & 46.2 & 73.0 & 82.7 & \textbf{2.0} & \textbf{8.7} \\
\midrule
\rowcolor{aliceblue!60} \textbf{Ours} & \textbf{48.4} & \textbf{74.6} & \textbf{83.7} & \textbf{2.0} & 9.0\\
\bottomrule[1.25pt]
\end{tabular}}
\caption{Comparisons to other methods on the Video-to-Text task on the MSVD, ActivityNet Captions, and DiDeMo datasets. ``$\uparrow$'' denotes higher is better, ``$\downarrow$'' denotes lower is better.}
\label{tab:comp_three_v2t}
\vspace{-0.5em}
\end{table*}

\subsection{Comparison with other Methods}
\noindent \textbf{Compared Methods.}\quad
We compare NeighborRetr with other text-image retrieval methods:
UNITER~\cite{chen2020uniter},
SOHO~\cite{huang2021seeing},
PCME~\cite{chun2021probabilistic},
IAIS~\cite{ren2021learning},
ViLT~\cite{kim2021vilt},
CSIC~\cite{liu2022image},
ViSTA~\cite{cheng2022vista},
ViLEM~\cite{chen2023vilem},
I-CLIP~\cite{dong2023giving},
CUSA~\cite{huang2024cross}.
We also compare NeighborRetr with other methods on the video-to-text retrieval task:
CLIP4Clip~{\cite{luo2021clip4clip}},
X-Clip~{\cite{ma2022x}},ECLIPSE~\cite{lin2022eclipse}
EMCL-Net~{\cite{jin2022expectation}}, HBI~\cite{jin2023video}, Diffusion~\cite{jin2023diffusionret}.

\noindent \textbf{Image-Text Retrieval.}\quad
Table~\ref{tab:itr_combined} presents a comprehensive comparison of our method against state-of-the-art approaches on the Flickr30K and MSCOCO datasets for both text-to-image and image-to-text retrieval tasks. On the Flickr30K, compared to the recent reinforcement learning-enhanced method I-CLIP~\cite{dong2023giving}, our approach shows a clear $1.5\%$ improvement on the text-to-image R@1 metric and a $5.2\%$ improvement on image-to-text R@1, even with I-CLIP’s additional steps to boost cross-modal alignment. On the MSCOCO dataset, compared to the soft-label supervision strategy of CUSA~\cite{huang2024cross}, which focuses on enhancing similarity recognition through uni-modal samples, our approach achieves a notable $2.1\%$ improvement on image-to-text R@1.

\noindent \textbf{Video-to-Text Retrieval.}\quad
We compare the proposed NeighborRetr method with other state-of-the-art methods on the Video-to-Text Retrieval task using the MSVD, DiDeMo, and ActivityNet Captions datasets, as shown in Table~\ref{tab:comp_three_v2t}. The results demonstrate that NeighborRetr outperforms the competing methods across all datasets. Specifically, in the MSVD dataset, NeighborRetr improves the R@1 score by $1.3\%$. For the DiDeMo dataset, NeighborRetr achieves an R@1 improvement of $2.2\%$ over HBI, indicating a significant advancement in the retrieval accuracy of long videos. These results consistently highlight the effectiveness of NeighborRetr in handling video-to-text retrieval tasks ranging from short clips to long videos.

\subsection{Ablation Studies}
\noindent \textbf{CLIP4Clip w/ NeighborRetr.}\quad
Integrating NeighborRetr into the CLIP4Clip~\cite{luo2022clip4clip} framework yields significant performance improvements, as evidenced in Table~\ref{tab:clip4clip}. Specifically, the R@1 score increases by $1.9\%$, the Rsum score improves by $4.7\%$, and MnR decreases by $2.2\%$. These metrics underscore the adaptability and robustness of NeighborRetr, demonstrating its potential for enhancing retrieval performance across different backbone architectures.

\begin{table}[t]
\centering
\small
\setlength{\tabcolsep}{1mm}{
\begin{tabular}{lccccc}
\toprule[1.25pt]
\multicolumn{6}{c}{MSR-VTT}\\
\cmidrule(rl){0-5}
Methods  & R@1$\uparrow$ & R@5$\uparrow$ & R@10$\uparrow$ & Rsum$\uparrow$ & MnR$\downarrow$ \\ \midrule
CLIP4Clip & 44.5 & 71.4 & 81.6 & 197.5 & 15.3\\
\midrule
\rowcolor{aliceblue!60} \textbf{w/ NeighborRetr} &  \textbf{46.4} & \textbf{73.0} & \textbf{82.8} & \textbf{202.2} & \textbf{13.1}\\
\bottomrule[1.25pt]
\end{tabular}}
\caption{Comparison of retrieval performance on the Text-to-Video task on MSR-VTT, demonstrating improved efficacy under the CLIP4Clip framework with our proposed NeighborRetr method.}
\label{tab:clip4clip}
\end{table}

\begin{table}[t]
\centering
\small
\setlength{\tabcolsep}{0.3mm}{
\begin{tabular}{cccccccccc}
\toprule[1.25pt]
$\mathcal{L}_{Wti}$ & $\mathcal{L}_{Opt}$ & $\mathcal{L}_{Nbi}$ & $\mathcal{L}_{KL}$ & skew$\downarrow$ & trunc$\downarrow$ & atkinson$\downarrow$ & robin$\downarrow$ & anti$\downarrow$ & hub$\downarrow$ \\ \midrule
\Checkmark & \ding{55} & \ding{55} & \ding{55} & 5.37 & 1.3 & 0.73 & 0.76 & 0.63 & 0.83 \\
\Checkmark & \Checkmark & \ding{55} & \ding{55} & 4.87 & 1.27 & 0.65 & 0.68 & 0.32 & 0.76 \\
\Checkmark & \ding{55} & \Checkmark & \ding{55} & 4.05 & 1.16 & 0.55 & 0.51 & 0.45 & 0.62 \\
\rowcolor{aliceblue!60}\Checkmark & \Checkmark & \Checkmark & \Checkmark & \textbf{3.20} & \textbf{1.04} & \textbf{0.44} & \textbf{0.45} & \textbf{0.23} & \textbf{0.58} \\ 
\bottomrule[1.25pt]
\end{tabular}}
\caption{Hubness metrics in the Text-to-Video retrieval task on the MSR-VTT dataset for all losses ablation. ``$\downarrow$'' denotes lower is better.}
\label{tab:hubness_of_diff_loss_comb}
\vspace{-1.0em}
\end{table}

\noindent\textbf{Hubness ablation of each loss.}\quad 
Table~\ref{tab:hubness_of_diff_loss_comb} demonstrates the effect of each loss function on hubness mitigation. The baseline model, without the proposed losses, exhibits the highest hubness values, indicating severe neighborhood relationship issues. Introducing $\mathcal{L}_{Wti}$ reduces \textit{skew} by $10\%$, boosting discriminative power by centralizing samples. Adding $\mathcal{L}_{Opt}$ significantly reduces hubness indicators, with \textit{anti} decreasing by $49\%$, in particular affects the uniform probability of anti-hub. Finally, including $\mathcal{L}_{Nbi}$ alongside the other losses yields the most substantial reductions: \textit{skew} drops by $17\%$, and \textit{hub} by $25\%$. These results indicate that $\mathcal{L}_{Nbi}$ is crucial in refining neighbor relationships and minimizing hub formation. Overall, the combined losses provide the most effective solution, balancing and diversifying neighbor interactions.

\begin{table}[t]
\centering
\small
\vspace{0.0em}
\setlength{\tabcolsep}{1mm}{
\begin{tabular}{ccccccccc}
\toprule[1.25pt]
Methods & skew$\downarrow$ & trunc$\downarrow$ & atkinson$\downarrow$ & robin$\downarrow$ & anti$\downarrow$ & hub$\downarrow$ \\ \midrule
w/o $\mathcal{L}_{Nbi}$ & 5.14 & 1.24 & 0.71 & 0.70 & 0.51 & 0.80\\
Simi & 4.47 & 1.17 & 0.50 & 0.66 & 0.37 & 0.72\\
Cent & 4.76 & 1.21 & 0.62 & 0.64 & 0.40 & 0.76\\
Simi+Cent & 4.12 & 1.15 & 0.54 & 0.60 & 0.30 & 0.70\\
\rowcolor{aliceblue!60} \textbf{Simi-Cent} & \textbf{3.20} & \textbf{1.04} & \textbf{0.44} & \textbf{0.45} & \textbf{0.23} & \textbf{0.58}\\
\bottomrule[1.25pt]
\end{tabular}}
\vspace{0.2em}
\caption{Hubness on the Text-to-Video task on MSR-VTT for $\mathcal{L}_{Nbi}$ loss ablation. ``$\downarrow$'' denotes lower is better.}
\label{tab:nighbor_loss_hub}
\vspace{-1.0em}
\end{table}

\noindent\textbf{Hubness ablation for formulations of $\mathcal{L}_{Nbi}$.}\quad 
In Table~\ref{tab:nighbor_loss_hub}, we analyze different formulations of cross-modal similarity (Simi) and centrality (Cent) for the hubness problem. The results show that incorporating Simi alone results in a modest skewness reduction of $13\%$, while Cent alone slightly decreases skewness by $7\%$, indicating the limitation of these formulations individually. The combined Simi+Cent formulation yields a certain decrease in skewness but remains suboptimal. However, the Simi-Cent formulation outperforms all other formulations, achieving a $38\%$ reduction in skewness, which highlights its effectiveness in reducing hubness across all metrics by better aligning the embedding space and fine-tuning more equitable neighborhood relations. This substantial improvement underscores the importance of the subtractive formulation in mitigating high-dimensional data biases.

\noindent\textbf{Test-Time Hubness Mitigation Analysis.}\quad 
Table~\ref{tab:infer_time} demonstrates that NeighborRetr achieves superior retrieval performance over the baseline while maintaining computational efficiency.
For training, NeighborRetr employs an online queue to maintain updated representations, mitigating staleness and removing outdated hubs. In inference, NeighborRetr is memorybank-free. Thus in Table~\ref{tab:infer_time}, NeighborRetr achieves 300× speedup than the QB-Norm method, which adopts a memory-bank as a test-time similarity normalization strategy.
Notably, NeighborRetr can improve R@1 by 2.7\% and R@Sum by 4.6\% with minimal inference time impact with the simplest similarity measure (Simi). The Simi-Cent variant enhances R@Sum by 0.2\% using only 1k validation samples compared to Simi. 
NeighborRetr effectively decouples representation quality from hubness bias during training, making test-time similarity adjustments redundant. This may explain why the test-time Simi-Cent metric does not translate to practical gains in NeighborRetr.
In contrast, QB-Norm methods require the entire training set (10k samples) and incur prohibitive inference costs. While both methods address the hubness problem, NeighborRetr operates during training for distribution-robust representations, whereas QB-Norm applies post-hoc similarity adjustments during inference, making their combination counterproductive (-2.5\% R@Sum) due to conflicting hubness mitigation strategies.

\begin{table}[t]
\centering
\scriptsize 
\vspace{0.6em}
\renewcommand{\arraystretch}{1.3}  
\setlength{\tabcolsep}{0.2mm}  
\begin{tabular}{l c c c c c c}
\toprule[1pt]
Method (Metric) & MB (Source/Size) & R@1$\uparrow$ & R@5$\uparrow$ & R@10$\uparrow$ & R@Sum$\uparrow$ & Inf. Time$\downarrow$ \\
\midrule
Baseline (Simi) & - & 46.8 & 73.6 & 82.7 & 203.1 & \textbf{19.30} \\
Ours (Simi) & - & \textbf{49.5} & \underline{74.1} & \underline{84.1} & \underline{207.7} & \underline{20.88} \\
Ours (Simi-Cent) & Val (1k) & \underline{49.1} & \textbf{74.5} & \textbf{84.3} & \textbf{207.9} & 23.37 \\
Baseline (+\textcolor{blue}{QB-Norm}) & Train (10k) & 47.7 & 74.0 & 83.7 & 205.4 & 5592.15 \\
Ours (+\textcolor{blue}{QB-Norm}) & Train (10k) & 48.6 & 73.8 & 82.8 & 205.2 & 5950.97 \\
\bottomrule[1pt]
\end{tabular}
\vspace{0.3em}
\caption{
Comparison of different similarity measures for Text-to-Video retrieval on MSR-VTT during inference, showing retrieval performance and inference time. ``$\uparrow$'' denotes higher is better, ``$\downarrow$'' denotes lower is better.
}
\label{tab:infer_time}
\vspace{-2.5em}
\end{table}

\begin{figure}[t]
\centering
\includegraphics[width=1.0\linewidth]
{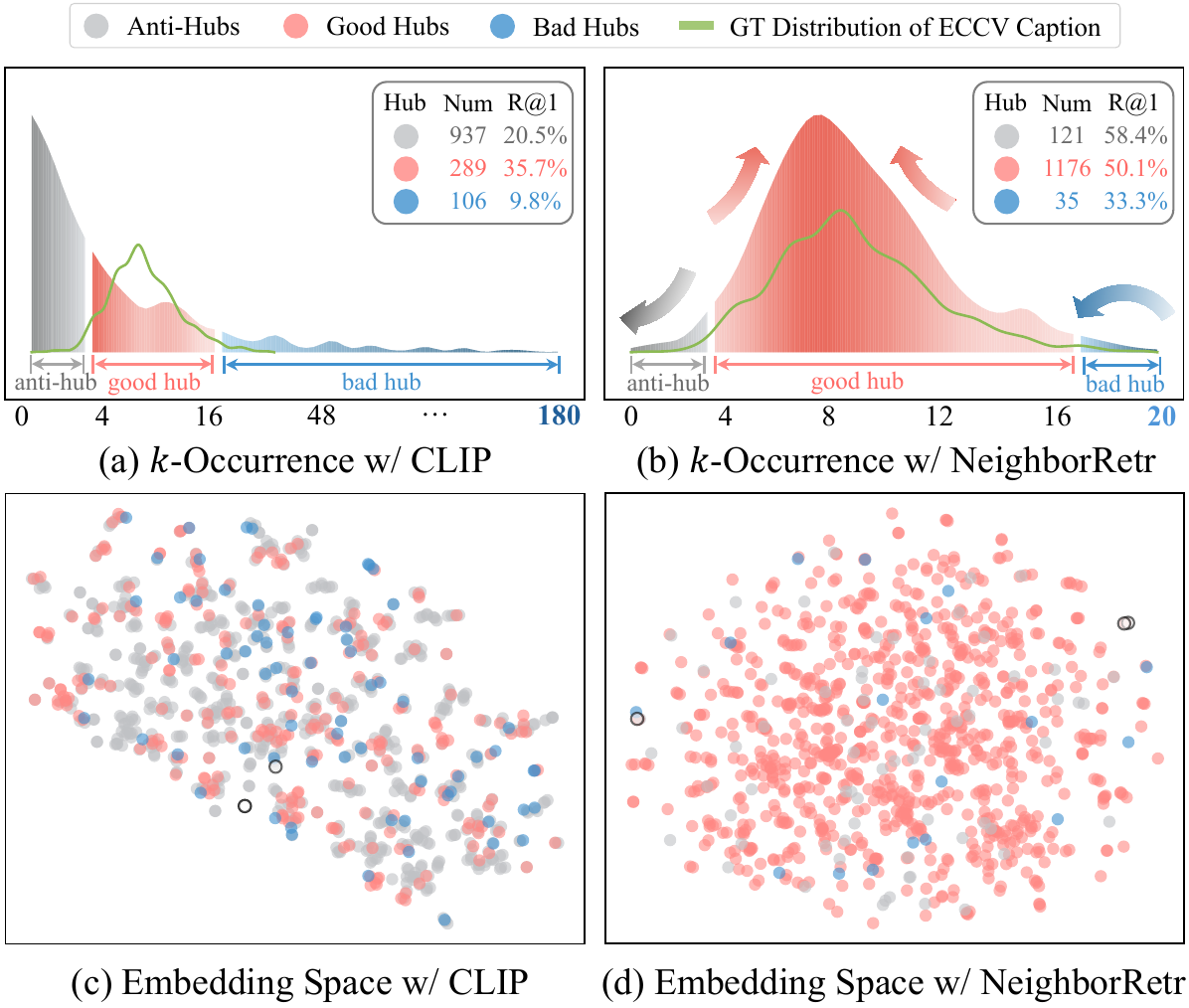}
\caption{
Visualization of hub frequency distribution and embedding space comparing CLIP vs. NeighborRetr, illustrating three hub types: \textbf{anti-hub $(k<4)$, good hub $(4\le k\le16)$, bad hub $(k>16)$}. NeighborRetr shows improved hub distribution and embedding characteristics.
}
\label{fig:hub_embedding}
\vspace{-1.2em}
\end{figure}

\begin{figure*}[t]
\centering
\includegraphics[width=1.0\linewidth]{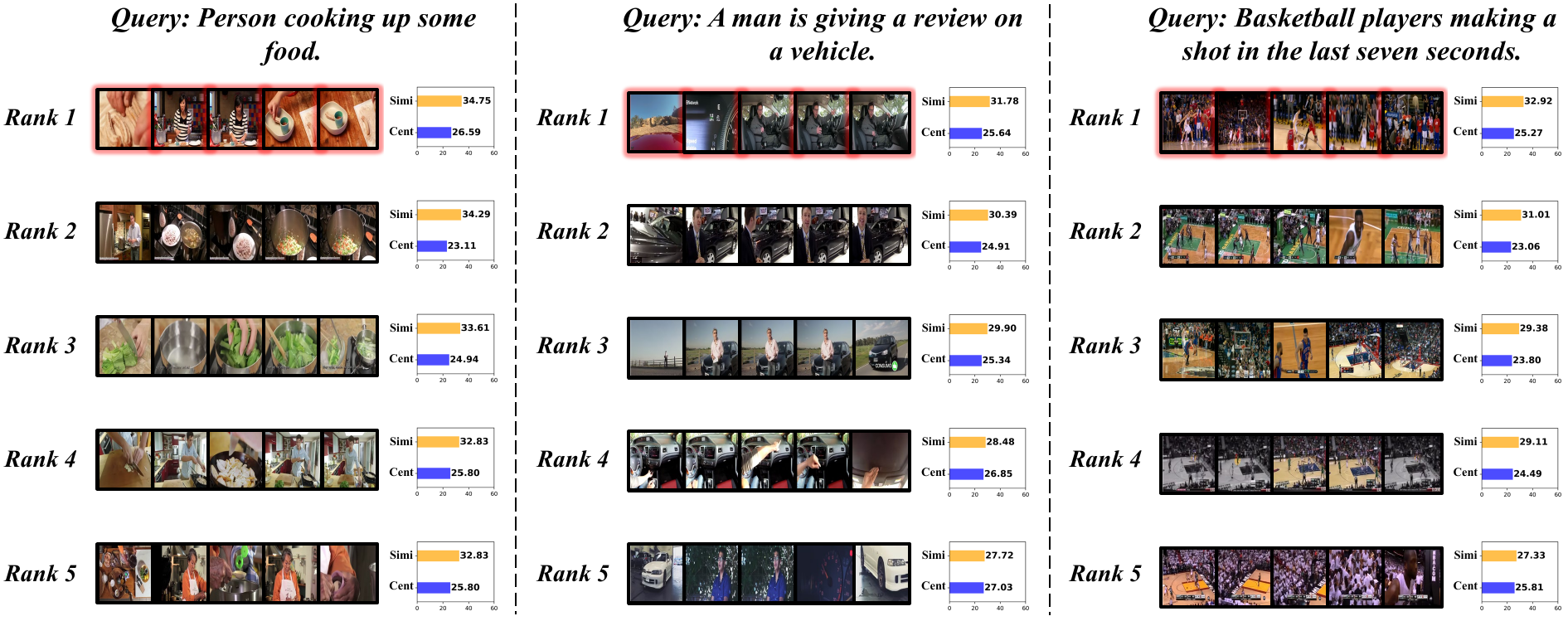}
\caption{Examples of text-to-video retrieval results with associated similarity (Simi) and centrality (Cent) scores. The ground truth video is highlighted in red. Each candidate video is assessed to demonstrate accurate text-to-video matching. For neighborhoods in Rank 2 to 5, NeighborRetr effectively identifies good neighbors, highlighting the effectiveness of our approach.}
\label{fig:supp_tog}
\vspace{-1.2em}
\end{figure*}

\subsection{Hub Distribution \& Embedding Space Analysis}
Fig.~\ref{fig:hub_embedding} illustrates hub distribution characteristics of CLIP versus NeighborRetr. We categorize hubs into three disjoint subsets: anti-hubs ($k<4$), good hubs ($4\leq k\leq16$), and bad hubs ($k>16$). NeighborRetr achieves a more balanced k-occurrence distribution, aligning more closely with the actual data distribution.
reducing isolated anti-hubs from 937 to 121, yielding a 37.9\% R@1 gain, and decreasing dominant bad hubs by 71, leading to a 23.5\% improvement. The embedding space shown in Fig.~\ref{fig:hub_embedding}(d) demonstrates a more uniform distribution, with good hubs gaining a 14.4\% improvement. This result confirms NeighborRetr effectively rebalances neighbor relationships by drawing relevant points closer while preventing irrelevant hubs from dominating neighborhoods.

\subsection{Visualization of Text-to-Video Retrieval}
Figure.~\ref{fig:supp_tog} demonstrates the effectiveness of our method in ranking videos that are highly relevant to the query while encompassing a range of semantically related scenarios. Higher-ranked videos show larger gaps between similarity (\textit{Simi}) and centrality (\textit{Cent}) scores, indicating the model's ability to prioritize less central samples, thereby reducing potential bias towards over-represented data. The lower-ranked videos showcase diversity, reflecting the model’s adaptability to various semantic contexts and its capability to match different topics within the same activity. As centrality scores increase with higher rankings, it becomes evident that our method not only captures good neighborhood relationships but also improves them dynamically, enhancing the overall retrieval quality.


\section{Hubness Metric Details}

\noindent\textbf{Definition of k-Occurrence.}\quad
In high-dimensional embedding spaces prevalent in cross-modal retrieval tasks, the metric of hubness manifests through certain samples' recurrent appearance as nearest neighbors. We quantify this phenomenon using the $k$-occurrence metric~\cite{tomavsev2014role}, $N_k(\mathbf{x})$, which indicates the frequency a sample $\mathbf{x}$ appears among the $k$-nearest neighbors of other samples:
\begin{equation}
N_{k}(\mathbf{x})=\sum_{i=1}^{n} p_{i, k}(\mathbf{x}),
\end{equation}
where $p_{i, k}(\mathbf{x})$ is a binary indicator function, defined as:
\[
p_{i, k}(\mathbf{x})=\left\{\begin{array}{ll}
1 & \text{if } \mathbf{x} \text{ is } k \text{ nearest neighbors of } q_{i}, \\
0 & \text{otherwise.}
\end{array}\right.
\]
This metric provides insights into the biases and centrality of samples within the embedding space, facilitating the analysis of hub formation and its impact on retrieval effectiveness.

\subsection{Distribution-based Metrics}
\noindent\textbf{Skewness.}\quad
We measure the asymmetry of the $k$-occurrence distribution using skewness ~\cite{radovanovic2010hubs}, denoted by $S_{N_k}$. It is calculated as:
\begin{equation}
S_{N_{k}} = \frac{E\left(N_{k} - \mu_{N_{k}}\right)^3}{\sigma_{N_{k}}^3},
\end{equation}
where $\mu_{N_{k}}$ and $\sigma_{N_{k}}$ are the mean and standard deviation of the $k$-occurrences, respectively. A high skewness value indicates a distribution with pronounced tails, suggesting the presence of hub samples that disproportionately appear as nearest neighbors, exacerbating the hubness issue.

\noindent\textbf{Skewness Truncated Normal.}\quad
The Skewness Truncated Normal~\cite{tomavsev2014role}, $S_{N_k}^{trunc}$, addresses skewness by truncating the $k$-occurrence distribution to enhance accuracy in central tendency analysis:
\begin{equation}
S_{N_k}^{trunc} = \frac{E\left[(N_k^{trunc} - \mu_{N_k^{trunc}})^3\right]}{(\sigma_{N_k^{trunc}})^3},
\end{equation}
where $N_k^{trunc}$ represents the adjusted $k$-occurrences, with $\mu_{N_k^{trunc}}$ and $\sigma_{N_k^{trunc}}$ detailing the mean and standard deviation of this truncated distribution, sharpening the focus on typical data behavior by excluding outliers.

\noindent\textbf{Atkinson Index.}\quad
The Atkinson Index~\cite{fischer2020unequal}, $A_{N_k}$, specifically measures the extent of inequality in $k$-occurrence distributions with a unique focus on the tails. This focus is modulated by the $\epsilon$ parameter, which provides a tunable sensitivity:
\begin{equation}
A_{N_k} = 1 - \frac{E[(N_k(x_i))^{1 - \epsilon}]^{\frac{1}{1 - \epsilon}}}{\mu_{N_k}},
\end{equation}
where $\epsilon$ enhances the index's responsiveness to the presence of hubs and antihubs,  crucial for nuanced analysis in cross-modal retrieval tasks. This adaptability allows for a tailored examination of how extremes centrality overall system fairness and data representation equity.

\noindent\textbf{Robin Hood Index.}\quad
The Robin Hood Index~\cite{feldbauer2018fast}, also referred to as the Hoover index, measures the inequality of the $k$-occurrence distribution. It reflects the extent of redistribution required to achieve uniformity in neighbor assignments across the dataset $D$:
\begin{equation}
\mathcal{H}^{k} = \frac{1}{2} \cdot \frac{E\left| N_{k} - \mu_{N_{k}} \right|}{\sum_{x \in D} N_{k}(x)},
\end{equation}
a higher index value signals greater inequality, which is indicative of significant disparities in the distribution of nearest neighbors, thereby highlighting potential issues in the fairness of retrieval system distributions.

\subsection{Occurrence-based Metrics}

\noindent\textbf{Antihub Occurrence.}\quad
Antihub~\cite{radovanovic2014reverse}, $A_{anti}$, identifies the proportion of samples that are not recognized as nearest neighbors by any other objects in the dataset:
\begin{equation}
A_{anti} = E[\mathbf{1}_{\{N_k(x_i) = 0\}}],
\end{equation}
where $N_k(x_i)$ is the $k$-occurrence of sample $x_i$, and $\mathbf{1}_{\{\cdot\}}$ indicates whether $x_i$ is completely ignored, emphasizing exclusion in the embedding space and spotlighting representational biases.

\noindent\textbf{Hub Occurrence.}\quad
Hub~\cite{radovanovic2010hubs}, $A_{hub}$, measures the prevalence of certain samples overly dominating nearest neighbor selections, affecting data representation equity:
\begin{equation}
A_{hub} = E[N_k(x_i) \cdot \mathbf{1}_{\{N_k(x_i) > k \cdot {\mathit{hub\_size}}\}}].
\end{equation}
where ${\mathit{hub\_size}}$ sets the threshold defining a hub if it appears frequently as a nearest neighbor. This metric, by quantifying the proportion of nearest neighbor slots occupied by hubs, highlights the potential skew in data representation.


\end{document}